\documentclass[12pt]{article}

\usepackage{newtxtext,newtxmath}

\usepackage{graphicx}

\usepackage[letterpaper,margin=1in]{geometry}
\usepackage{tabularx}
\usepackage{booktabs}
\usepackage{subcaption}
\usepackage{amsmath}
\usepackage{longtable}
\usepackage{geometry}
\usepackage{multirow}
\usepackage{caption}
\usepackage{array}
\usepackage{ragged2e}
\renewenvironment{abstract}
	{\quotation}
	{\endquotation}

\date{}

\makeatletter
\renewcommand{\fnum@figure}{\textbf{Figure \thefigure}}
\renewcommand{\fnum@table}{\textbf{Table \thetable}}
\makeatother

\usepackage{scicite}

\usepackage{url}

\def\scititle{Gendered Divides in Online Discussions about Reproductive Rights}
\title{\bfseries \boldmath \scititle}

\author{
    Ashwin~Rao$^{1,2\ast}$,
    Sze~Yuh~Nina~Wang$^{3}$,
    Kristina~Lerman$^{4}$\and
    \small$^{1}$Thomas Lord Department of Computer Science, University of Southern California, Los Angeles, CA, USA.\and
    \small$^{2}$Information Sciences Institute, University of Southern California, Marina del Rey, CA, USA.\and
    \small$^{3}$Department of Psychology, Cornell University, Ithaca, NY, USA.\and
    \small$^{4}$Luddy School of Informatics, Indiana University, Bloomington, IN, USA.\and
    \small$^\ast$Corresponding author. Email: ashwinra@usc.edu\and
}

\begin{document}

\maketitle

\begin{abstract} \bfseries \boldmath
The U.S. Supreme Court’s 2022 ruling in \textit{Dobbs v. Jackson Women’s Health Organization} marked a turning point in the national debate over reproductive rights. While the ideological divide over abortion is well documented, less is known about how gender and local sociopolitical contexts interact to shape public discourse. Drawing on nearly 10 million abortion-related posts on X (formerly Twitter) from users with inferred gender, ideology and location,  we show that gender significantly moderates abortion attitudes and emotional expression,  particularly in conservative regions, and independently of ideology. This creates a gender gap in abortion attitudes that grows more pronounced in conservative regions. The leak of the \textit{Dobbs} draft opinion further intensified online engagement, disproportionately mobilizing pro-abortion women in areas where access was under threat. These findings reveal that abortion discourse is not only ideologically polarized but also deeply structured by gender and place, highlighting the central the role of identity in shaping political expression during moments of institutional disruption.
\end{abstract}

Few issues have divided the American public as deeply---or as enduringly---as reproductive rights. Long a flashpoint in cultural and political battles, abortion debates have come to symbolize broader struggles over bodily autonomy, religious freedom, and gender equality. The 2022 Supreme Court ruling in \textit{Dobbs v. Jackson Women’s Health Organization}, which overturned nearly five decades of federal protections for abortion access established by \textit{Roe v. Wade}, marked a seismic shift. It not only intensified existing partisan divides~\cite{rao2025polarized,levendusky2024has}, but also reshaped the legal and political terrain, triggering abrupt policy reversals in many states and catalyzing a realignment in the national debate over reproductive rights.

A growing body of research has documented partisan cleavages in public attitudes toward reproductive rights~\cite{gallup2010abortion,gallup2011abortion,gallup2021abortion,dimaggio1996have,evans2003have,rao2025polarized}. However, less attention has been paid to the way in which gender and sociopolitical environment jointly shape both opinion formation and patterns of public expression. Recent surveys point to a widening gender gap in political orientation, particularly among younger voters. For example, in the 2024 U.S. presidential election,  white men predominantly supported President Trump, while white women preferred Vice President Harris~\cite{murray2024nbc}. Similarly, Gallup polling found a sharp increase in the share of young women identifying as politically liberal and supporting reproductive rights~\cite{saad2024gallup}. While women consistently report higher support for abortion access, particularly in countries with less restrictive policy environments~\cite{loll2019differences,gallup2023abortion}, men, even those who identify as pro-choice, often show less engagement with the issue~\cite{americansurveycenter2023abortion,gallup2023abortion,lizotte2015abortion}. 

Prior work has also documented gendered modes of engagement in online discourse around reproductive rights~\cite{rao2025polarized,levendusky2024has}. Men are more likely to frame abortion in moral terms aligned with anti-abortion positions~\cite{zhang2016gender}, while women are more likely to engage through lower-visibility actions such as retweeting, liking, and commenting~\cite{bode2017closing,hu2023tweeting,philippe2024abortion}. These differences highlight the importance of examining how gender shapes not only the content of political attitudes but also the mode of political engagement.

Additionally, there is increasing recognition that political expression is shaped by structural and environmental factors. Prior work has shown that local policy regimes, economic conditions, and access to reproductive healthcare influence a range of health and social outcomes~\cite{braveman2011social,dawes2020political,jaidka2021information}, yet their role in structuring online political discourse remains insufficiently understood. Although recent research has began to link state-level sociopolitical indicators to aggregate abortion attitudes~\cite{aleksandric2024analyzing}, it is unclear how such contexts asymmetrically shape the expressive behaviors of men and women. These questions have gained new urgency in the post-\textit{Dobbs} era, as legal and political authority over reproductive rights has shifted decisively to the state level, intensifying the salience of local context in shaping public engagement.

To investigate these questions, we analyze millions of abortion-related tweets posted throughout 2022~\cite{chang2023roeoverturned}.
Leveraging state-of-the-art natural language processing methods, we infer user gender~\cite{van2023open}, ideology~\cite{barbera2015birds,nettasinghe2025group}, stance toward abortion, emotional language~\cite{alhuzali2021spanemo,rao2025polarized}, and geographic location~\cite{dredze2013carmen}, giving us with almost 10 million tweets from 805 thousand users with known gender and location. 
The spatial and temporal granularity  of our data allows us to integrate it with county- and state-level structural indicators---including Republican vote share, income, housing costs---to explore how regional sociopolitical environments shape online abortion discourse. We further investigate how key events, like the leak of the \textit{Dobbs} decision, influenced patterns of political. We examine not only gender differences in political expression but also variation in emotional tone and online engagement across sociopolitical contexts.

This study makes three primary contributions. First, we identify a consistent gender gap in support for abortion rights, with women expressing stronger support than men---a divide that is especially pronounced in ideologically conservative states. Second, we uncover systematic gender differences in emotional expression: men are more likely to express anger and disgust, while women express more fear, sadness, and---interestingly---optimism. These emotional differences are magnified in conservative states, particularly among pro-choice women. 
Third, we show that the leak of the \textit{Dobbs} decision mobilized pro-choice engagement not just in liberal states, but also in conservative regions, largely driven by newly active women. This response contributed to a widening gender gap in abortion discourse.

By applying computational methods to large-scale social media data, our study provides a dynamic, fine-grained account of how gender, ideology, and sociopolitical context shape public engagement with reproductive rights in the post-\textit{Dobbs} era.

\section{Results}
We analyze 10 million tweets from 805 thousand users who posted about abortion on Twitter in 2022~\cite{chang2023roeoverturned} from a location in the U.S. and whose gender we were able to infer (see Methods).
Descriptive statistics are provided in Table \ref{tab:user_tweet_summary} in SI Section \ref{sec:additional} of Supplementary Information. 

\begin{figure*}[!ht]
    \centering
    \includegraphics[width=0.9\linewidth]{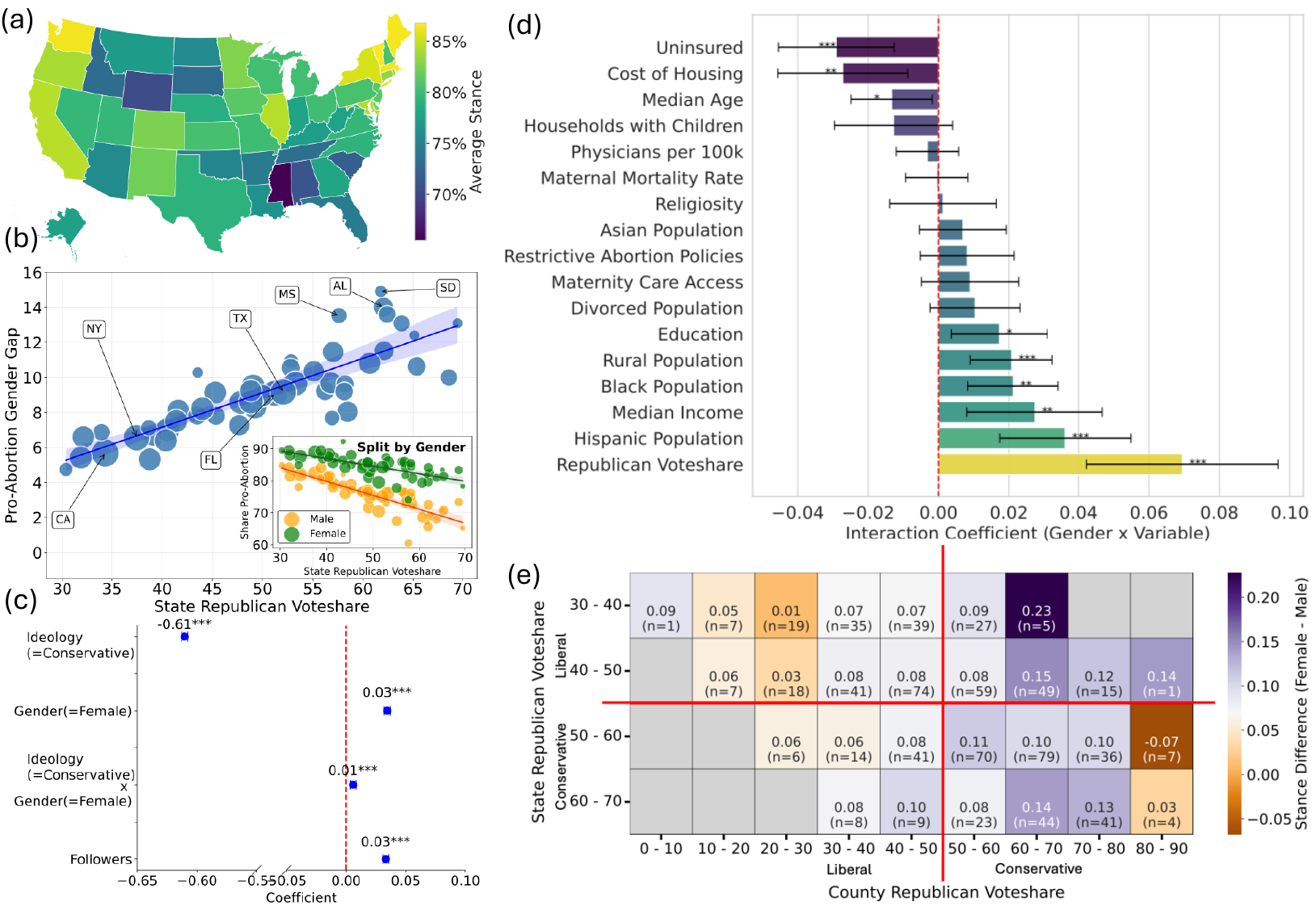}
    \caption{\textbf{Gender Gap in Support for Reproductive Rights.} 
    (a) State-level proportion of pro-abortion tweets among all tweets expressing a clear stance. (b) Gender gap in support for reproductive rights increases with Republican voteshare at the state level. The inset shows that this divergence in conservative states is primarily driven by men’s lower pro-abortion sentiment. 
    (c) Results from a mixed-effects regression predicting abortion stance from user-level ideology and gender. Even after controlling for ideology, gender remains a statistically significant predictor of stance. 
    (d) Socio-demographic correlates of the gender gap in abortion stance. Republican vote share, percentage of Black and Hispanic populations, median income, percentage of rural residents, median age, housing costs, and percentage of uninsured individuals are statistically significant predictors of the gender gap in abortion attitudes. Asterisks indicate statistical significance ($\ast p < 0.05$, $\ast\ast p < 0.01$, $\ast\ast\ast p < 0.001$). (e) Gender differences in political stance by state and county partisan context. Heatmap shows the difference in average stance between women and men (women - men) across state and county Republican voteshare bins (10\% increments). Number of counties in each state-county voteshare bin is in parenthesis.
    }
    \label{fig:gender_gap}
\end{figure*}

\paragraph{Gender Gap in Pro-Abortion Attitudes}

Attitudes toward abortion vary widely across the United States. Figure~\ref{fig:gender_gap}(a) shows the share of pro-abortion tweets among all abortion-related posts expressing a clear stance, aggregated by user state.  Regional differences are pronounced: users in the Northeastern and Western states express more support for abortion rights, while states across the South and Midwest express less support. These patterns mirror state-level policy differences: states with more restrictive abortion laws correspond to lower online pro-abortion sentiment. 

Support for abortion rights declines sharply in more conservative states, as measured by Republican vote share in the 2020 presidential election, and this decline is steeper among men than women. Consequently, the gender gap in pro-abortion sentiment widens in conservative states (Pearson’s $r = 0.83$, $p < 0.001$, Figure~\ref{fig:gender_gap}(b)).

To disentangle the contributions of individual- and environment-level factors to these gender gaps, we proceed in two steps.
First, we assess the impact of individual-level characteristics on support for abortion rights using a regression model that includes user gender, political ideology, and their interaction (Figure~\ref{fig:gender_gap}(c)). While ideology is a strong predictor---conservative users are significantly less likely to support abortion rights ($\beta = -0.61$, ***$p < 0.001$)---gender retains a significant and independent effect:  women are more supportive of abortion rights than men ($\beta = 0.03$, ***$p < 0.001$). Notably, the positive interaction term suggests that conservative women are  more supportive of abortion than their male ideological counterparts ($\beta = 0.01$, ***$p < 0.001$). This within-group heterogeneity---especially among conservatives---demonstrates the importance of identity in conditioning ideological expression. Results of this regression are shown in Table \ref{tab:mixedlm_ideo} in  Section~\ref{sec:additional} of Supplementary Information.

Next, we explore how local sociopolitical context at the state level influences these patterns. 
Using a mixed effects linear model with  state as a random effect,  we estimate interaction effects between user gender and key sociopolitical indicators, including economic vulnerability, healthcare access, family structure, education, racial composition, and ideological climate (Fig.~\ref{fig:gender_gap}(d)). These interaction terms capture how specific features of local environment differentially shape abortion attitudes by gender. 

For example, after controlling for ideology, {economically vulnerable} environments---states with more {uninsured} or higher {housing costs}---lower women's support for reproductive rights relative to men, while more educated, racially diverse, and affluent states tend to exhibit larger gender gaps.
 
These results highlight that political expression is not only shaped by individual ideology and gender but also by the broader sociopolitical context in which individuals are embedded.

These patterns also persist at finer geographic resolution. For robustness, we analyze the gender gap in abortion attitudes at the county level (Supplementary Fig.~\ref{fig:gender_gap_county} (a)--(d)).  This additional analysis confirms that the gender gap widens in more conservative counties. Interactions effects analysis shows that the gender gap is correlated with largely similar county-level variables (SI Fig.~\ref{fig:gender_gap_county}(e)).

To further local context, we plot the gender gap across counties, grouped by both county-level and state-level Republican vote share (Figure~\ref{fig:gender_gap}(e)). The x-axis bins counties by local Republican support, while the y-axis categorizes them by the broader state-level political climate. Each cell reflects the median gender gap in counties that fall into the respective bin, with darker purple shades indicating larger gaps favoring pro-abortion sentiment among women, while lighter and orange shades represent smaller or reversed gaps.
The most pronounced gender gaps appear in conservative counties (particularly in the 60–70\% Republican vote share) located within liberal states ($<50\%$ Republican vote share). This suggests that local political environments---not just state-level partisanship---play an important role in shaping gendered political expression.

Together, these analyses reveal that gender significantly shapes attitudes toward abortion beyond ideological affiliation, and that local sociopolitical conditions differentially affect how men and women express support for reproductive rights.

\begin{figure*}[!ht]
    \centering
    \includegraphics[width=0.9\linewidth]{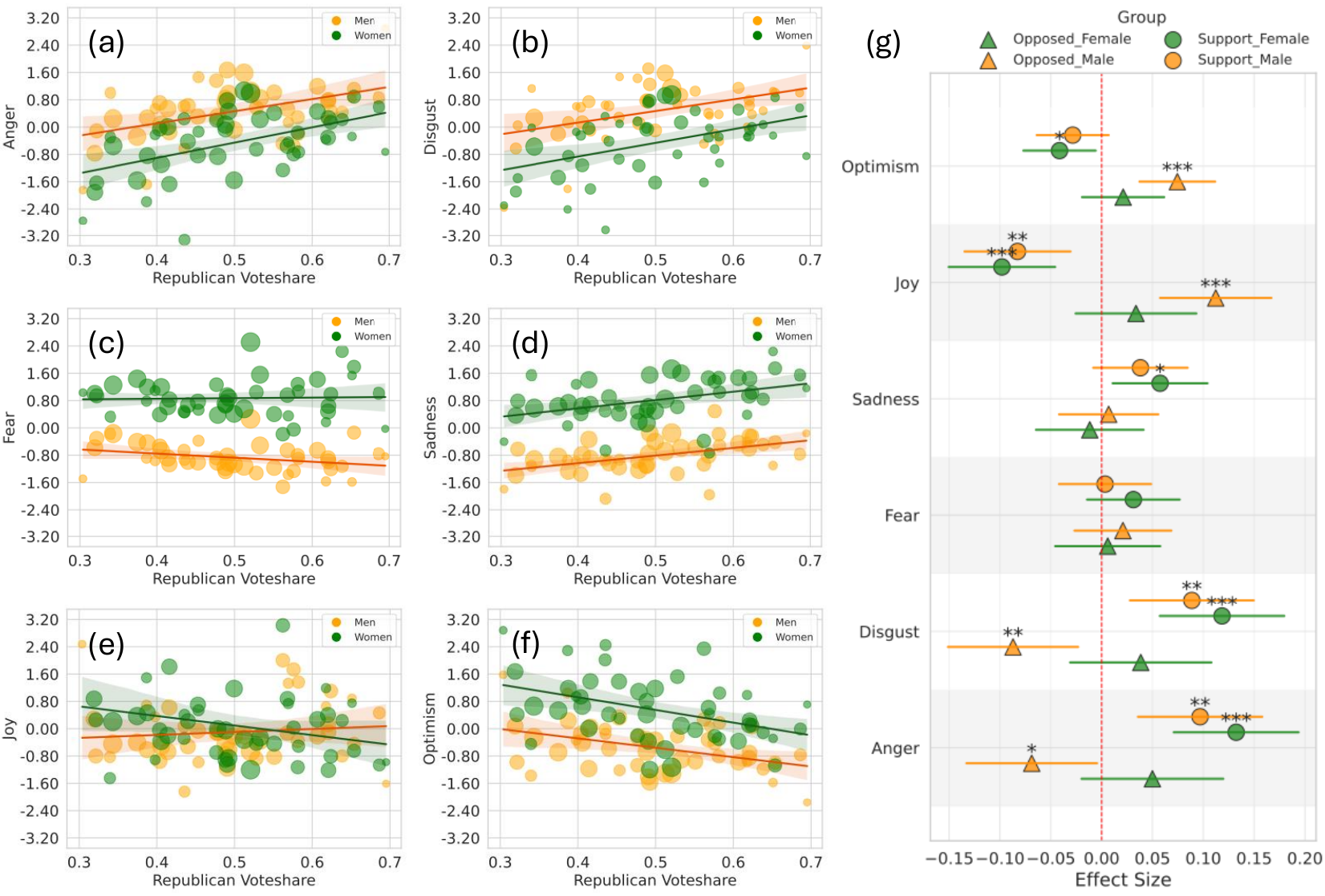}
    \caption{\textbf{Gendered Differences in Emotional Expression Across Political Contexts.} (a-f) shows gendered patterns in the expression of six core emotions---anger (a), disgust (b), fear (c), sadness (d), joy (e), and optimism (f)---as a function of Republican vote share across U.S. states. We use z-score standardization on the y-axes in (a-f) for consistent scales across emotions. Panel (g) presents group-specific effects from a mixed-effects regression modeling emotional expression as a function of state ideology and user group (gender and abortion stance). Results reveal divergent emotional patterns, with women—especially those supportive of abortion access—expressing more anger, disgust, sadness, and less joy in conservative contexts, while men opposing abortion access show increased positive emotions.}

    \label{fig:emotion_gap}
\end{figure*}

To assess whether the observed gender gap in abortion stance reflects broader gendered patterns in polarized online discourse, we compare it to attitudes toward mask-wearing during the COVID-19 pandemic. Using tweets about the pandemic posted in 2020 and 2021, we identity posts about masking, an issue that became ideologically polarized~\cite{rao2023pandemic} but was not strongly associated with gender identity.  Across the full sample of users with known gender and location, we observe a gender gap in pro-masking attitudes, which increases in the more conservative states (see SI Fig.~\ref{fig:masking_gap}). However, this gender gap (mean $=0.035 \pm 0.016$, median $=0.046$) is just a fraction of the  gender gap in pro-abortion attitudes (mean $=0.091 \pm 0.022$, median $=0.093$).   
Restricting the analysis to 262,852 users active in both abortion and masking discussions results in a much bigger gender gap for the abortion issue (mean $=0.056 \pm 0.023$, median $=0.061$) compared to masking (mean $=0.030 \pm 0.023$, median $=0.036$).
These findings suggest that gendered political expression is especially pronounced in the context of reproductive rights. Full results and summary statistics are provided in Supplementary Information Section~\ref{sec:masking} and Table~\ref{tab:gender_gap_summary}.

\paragraph{Emotional Differences in Abortion Discussions}

Ideology plays a major role in shaping the emotional dynamics of political discourse, especially around contentious issues like abortion~\cite{iyengar2015fear,nettasinghe2025out}. 
However, the role of gender in the emotional divides of polarized debates has been underexplored.
Prior work has shown that men and women differ in engagement strategies and emotional expression  in online political speech~\cite{bode2017closing,hu2023tweeting,philippe2024abortion}, yet these patterns have rarely been analyzed in relation to contentious issues like reproductive rights. To address this gap, we examine how emotional expression in abortion-related discourse on Twitter varies by gender and political context.  

Our analysis focuses on six key emotions: anger, disgust, fear, sadness, joy, and optimism. Figure~\ref{fig:emotion_gap} presents emotion prevalence in the discussion of reproduction rights across states, conditioned on gender and state Republican vote share. Several clear patterns emerge. Negative emotions, like anger (panel a), disgust (panel b), and sadness (panel d) increase as states become more conservative. Fear, however, remains relatively flat across political contexts. Men consistently express more anger and disgust, emotions often linked to moral outrage~\cite{brady2021social}, while women express more fear and sadness. 

Positive emotions show contrasting trends. Optimism (panel f) declines in the more conservative states for both genders. Joy (panel e) displays an interesting crossover: as Republican voteshare increases, men's expression of joy rises slightly, while women's declines. While women's optimism is consistently higher across political contexts, they express more joy than men only in the least conservative states. 

To better understand how the state's political climate affects shapes the emotional tone of online discussions, we model user emotions as a function of state-level political context, user gender and abortion stance. We infer user-level stance by computing a weighted average of each user's tweet-level stances and binarize this measure at a threshold of 0.5. 
Using mixed effects regression with state as a random effect, we estimate interaction effects between state's Republican vote share and user group membership (defined by gender and stance on abortion rights). Figure \ref{fig:emotion_gap}(g) displays the estimated changes in emotional expression across user groups as states become more ideologically conservative. 

We find that, as states become more conservative, users who support abortion rights, particularly women, express significantly more anger and disgust.
In contrast, men who oppose abortion express less anger and disgust as their state becomes more conservative. 
Men who oppose abortion also show significantly more joy and optimism the more conservative states, whereas users supporting abortion rights, especially women, show a markedly less joy and more sadness in the more conservative contexts. Fear, however, does not exhibit statistically significant changes across any group.
The results of this regression are detailed in Table \ref{tab:emotion_regressions} under SI Section \ref{sec:additional}.

These findings suggest emotional expression is shaped not only by users’ stance on abortion and the political climate,  but also strongly modulated by gender, reflecting different stakes in the abortion debate. Notably, men and women who support abortion rights exhibit broadly similar emotional responses across political contexts: expressing more anger and disgust and less joy in the more conservative contexts, where reproductive rights are more restricted. In contrast, among those who oppose abortion rights, only men’s emotional expressions vary with state conservatism, becoming more positive in ideologically aligned states. The emotional expressions of women who oppose abortion remain largely unaffected by the political climate.

\paragraph{The Effect of SCOTUS's Leaked Ruling}

\begin{figure*}[!ht]
    \centering
    \includegraphics[width=0.7
    \linewidth]{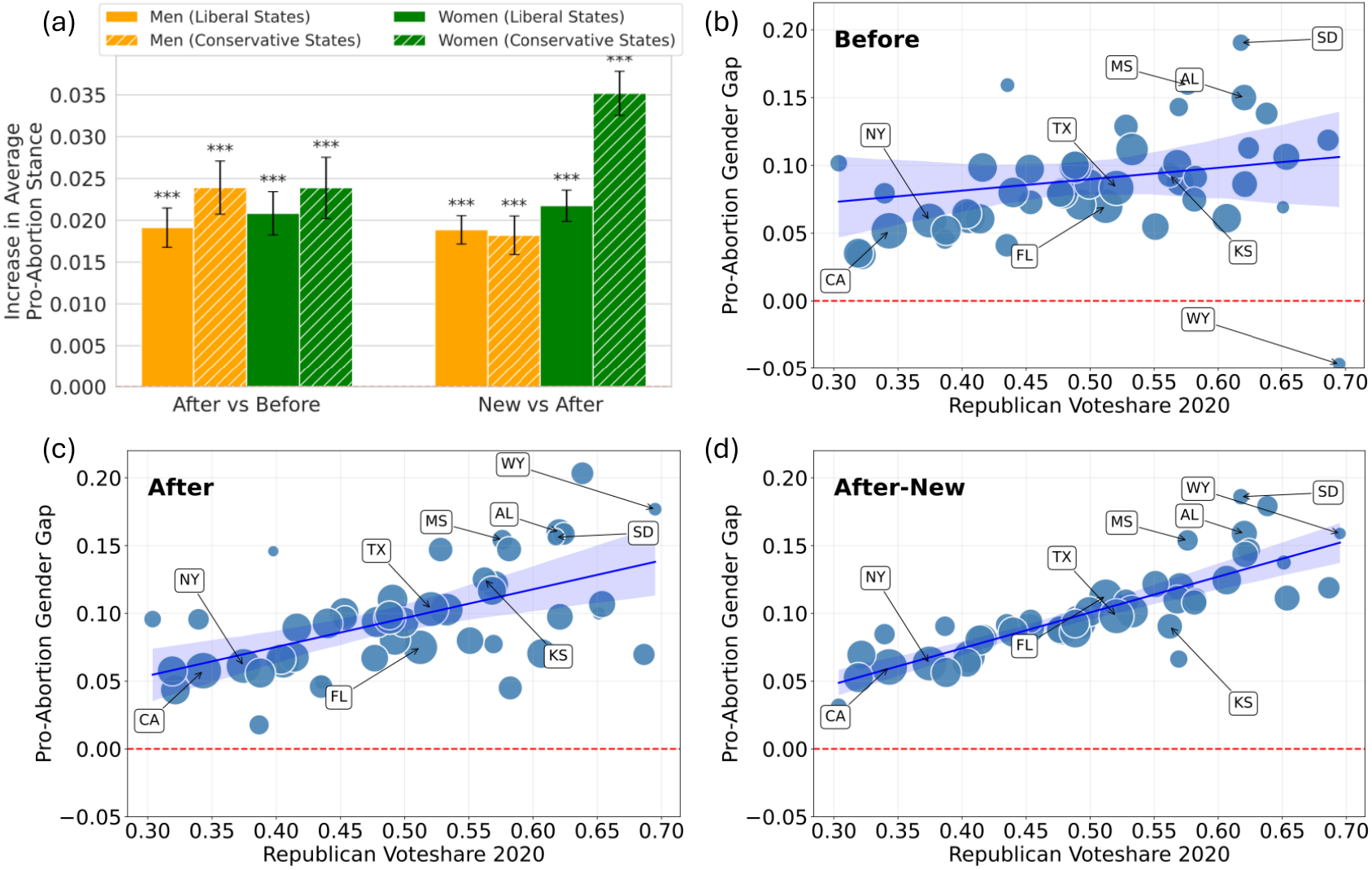}
    \caption{\textbf{Impact of the Dobbs Leak on Abortion Attitudes and Gender Polarization.}  This figure examines shifts in abortion stance following the May 3, 2022, leak of the Dobbs draft opinion. Panel (a) shows within-subject changes among users active both before and after the leak, alongside differences between pre-leak and newly active users. The largest increases in support occurred among users in conservative states—particularly women. Comparing pre-leak and new users reveals that newcomers, especially women in conservative states, expressed significantly stronger pro-abortion views. Panels (b)--(d) display the gender gap in abortion stance by state: before the leak (b), after the leak (c), and among new users who joined post-leak (d). The gender gap widened substantially after the leak for pre-leak users, especially in more conservative states, and new post-leak users (d) also exhibited wider gender gaps in these contexts.}
\label{fig:stance_before_after}
\end{figure*}

The publication on May 3, 2022 of the leaked draft opinion in Dobbs v. Jackson Women’s Health Organization~\cite{gerstein2022supreme}, marked a pivotal moment in the national conversation on abortion rights. Written by Justice Samuel Alito, the opinion proposed overturning Roe v. Wade, thereby eliminating federal protections for abortion access. This unprecedented breach inflamed public  discourse on both sides of the political spectrum~\cite{rao2025polarized}. We analyze how online behaviors changed in the aftermath of the leak to understand the event’s broader impacts on public opinion. Specifically, we examine how stances of users who were active both before and after the leak may have changed, and compare them to users who entered the conversation after the leak. By analyzing activity patterns, stances, and the geographic distribution of new users, we gain insight into how political events mobilize segments of the population. 

To measure the impact of the leak, we employed a linear regression model incorporating a three-way interaction among user gender, their state's political leaning, and the time period (pre- vs.\ post-leak). We classified a state as liberal if more than 50\% of its votes in the 2020 Presidential Election went to the Democratic candidate, and conservative otherwise. Two complementary analyses were conducted:
\begin{enumerate}
    \item \textbf{Within-Subject Analysis:} We examined approximately 300{,}000 users who were active (i.e., posted at least once) during both the pre- and post-leak periods to evaluate individual-level changes in abortion stance. The pre-leak period spanned 122 days, from January 1, 2022, to May 2, 2022, while the post-leak period encompassed the subsequent 122 days, from May 3, 2022, to September 3, 2022.
    
    \item \textbf{Between-Group Comparison:} We compared users who were active before the leak (from January 1, 2022, onward) with those who began participating only after the leak, to identify differences in stance between early participants and newcomers to the conversation.
\end{enumerate}

These complementary analyses allow us to assess both within-individual changes in stance among users already active in online abortion discussions and shifts due to newly mobilized participants. According to within-subject analysis (Refer After vs Before Fig.~\ref{fig:stance_before_after}), both men and women in conservative states registered a larger increase in  support for abortion rights after the leak---approximately 2–3 percentage points---compared to users in liberal states.

The between-group analysis reveals that women from conservative states who joined the online conversation after the leak were on average 3.5\% more supportive of abortion rights than women from  conservative states who were active before the leak (Refer New vs After Fig.~\ref{fig:stance_before_after}). A similar, though smaller, pattern exists among newly active men from conservative states, as well as among newly active users from liberal states. These findings suggest that while the leak had only a modest effect on the stances of users already engaged in abortion discussions, it mobilized a new cohort of participants who were more supportive of abortion rights. The disproportionate increase in in pro-abortion sentiment---particularly among women in conservative states---indicates a potential counter-reaction by those who viewed the leaked opinion as a direct threat to their rights and autonomy. Regression results are shown in Tables \ref{tab:stance_before_after} and \ref{tab:stance_before_after_new} in Section \ref{sec:additional} of SI.   

Figures~\ref{fig:stance_before_after}(b)--(d) illustrate the gender gap in abortion attitudes for the approximately 300K users active both before and after the leak, and new users who joined post-leak. The gender gap (i.e., the difference in the share of pro-abortion tweets between women and men) was modestly correlated with Republican vote share before the leak (Fig.~\ref{fig:stance_before_after}(b), Pearson $r = 0.20$, regression coefficient $\beta= 0.08$), but increased substantially among the same users after the leak (Fig.~\ref{fig:stance_before_after}(c), $r = 0.57$, $\beta =0.21$). The strongest effect appears among new users, with correlation of $r = 0.84$ (Fig.~\ref{fig:stance_before_after}(d), $\beta= 0.27$), suggesting that the leak of the Dobbs decision increased gender polarization around abortion.  Additionally, we analyze how engagement from these groups evolved pre- and post-leak in SI Section \ref{sec:additional}.

We attribute the widening gender gap in abortion attitudes following the Dobbs leak to two dynamics: greater activity among existing pro-abortion users after the leak, and the influx of newly mobilized users, particularly pro-abortion women in conservative states (see Supplementary Information Fig.~\ref{fig:activity_change} and Section~\ref{fig:activity_change}).

\section{Discussion}
This study provides new evidence about the gendered dynamics of abortion discourse, using social media data to reveal how users' social and political environments  and identity jointly shape their online expression, including their  stances on reproductive rights and mobilization in the wake of a major legal disruption.  While ideology remains the strongest determinant of political expression, we find that gender serves as a critical  moderator. Women, especially those in conservative states, exhibit distinct patterns of abortion support, affect, and engagement that diverge from both male counterparts and from prevailing partisan norms in their communities.

These dynamics contribute to a gender gap in abortion attitudes, with women expressing greater support for reproductive rights than men---a gap that widens in more conservative regions. This divide became even more pronounced following the leak of the Dobbs v. Jackson Women’s Health Organization decision, which served as a flashpoint that mobilized a new wave of users.Notably, these newly engaged voices were disproportionately pro-choice, especially in regions where abortion access was most threatened.

In addition, we find that emotional expression around abortion is deeply gendered and sensitive to political context. First, there is a consistent gender gap in emotion: women express more fear, sadness, and optimism, while men express more anger and disgust, regardless of ideological environment. Second, negative emotions intensify in conservative states for both genders, with higher levels of anger, disgust, and sadness and declining optimism. Interestingly, while pro-abortion users of both genders display similar emotional patterns, anti-abortion men differ not only from abortion-rights supporters but also from anti-abortion women, suggesting that emotional reactions to the issue are shaped by both gender and stance in asymmetrical ways.

While this study provides valuable insights into gender dynamics within abortion-related discourse on social media, several limitations warrant consideration. First, we use a name-based gender classifier to categorize users as either \textit{Man} or \textit{Woman}. Though suitable for large-scale analysis, this method may misgender individuals or exclude non-binary users, thereby limiting the inclusivity of our findings. Our analysis focuses on the two largest gender groups, but future research should adopt more inclusive classification techniques to better capture the diversity of gender identities. Second, although we employ LLMs to classify stance and ideology---with validation indicating strong overall performance---these models are not infallible. In particular, emotion classification remains a challenging and inherently subjective task, and performance metrics reflect this difficulty. Future work may benefit from fine-tuning LLMs on domain-specific data to improve accuracy in detecting nuanced affective expressions. Third, we acknowledge that Twitter (now X) does not represent the full spectrum of public opinion in the United States. Its user base is skewed by ideology, age, and geography. A more comprehensive understanding of abortion discourse would benefit from incorporating data from additional platforms, such as Reddit, Facebook, TikTok, Gab, and Parler, where distinct communities may engage with the topic in different ways. We selected Twitter for its central role in political communication~\cite{pew2022partisan}, its influence on political agendas~\cite{barbera2019leads}, its impact on news media coverage~\cite{hanusch2019comments,nelson2019doing}, and its use as a proxy for public opinion by journalists and campaign strategists~\cite{mcgregor2019social,mcgregor2020taking}. Additionally, we are unable to draw any causal or directional claims from the current analyses (e.g., whether user's expressions are shaped by their sociopolitical contexts, or whether they select into these contexts based on their attitudes).  Finally, our geolocation approach is heuristic-based and relies on users' self-reported profiles, which may introduce inaccuracies. Future research could employ more robust location inference methods or pair online discourse with offline demographic data to improve the reliability of geographic analyses. While our study yields valuable insights, it is confined to abortion-related discourse from a single year and within one nation's context. To evaluate the broader applicability of our findings, future research should replicate this analysis across multiple time periods and in other countries where abortion remains a contentious issue. Nevertheless, the large scale of our dataset, rigorous validation procedures, and aggregation of results help to reduce the potential impact of these limitations.

Taken together, our findings show that the abortion debate on social media is not only ideologically polarized but also polarized by gender. The Dobbs leak expanded these divides, not merely by increasing disagreement, but by altering who participates, how frequently, and with what emotional tone. In doing so, it transformed the online abortion conversation, particularly in conservative states, into a space increasingly dominated by pro-choice voices.

These insights have broader implications for understanding political discourse and mobilization in digital spaces. They suggest that major policy disruptions do more than shift opinions---they reshape emotional landscapes and create new identity-driven fault lines in already polarized debates. Future research should examine whether similar patterns emerge in other domains and how platform dynamics mediate the amplification and persistence of emotionally charged political expression.

\section{Methods and Data}
\subsection{Twitter Data}
We analyze a dataset of tweets about abortion rights collected from January 1, 2022, to January 1, 2023~\cite{chang2023roeoverturned}. The dataset comprises 57,540,676 tweets from 5,426,555 users, gathered using keywords and hashtags related to abortion discussions in the United States. Of these, 35,949,368 tweets from 2,877,409 users include inferred gender information, and 9,982,073  tweets from 805,382 users include both inferred gender and U.S. state location (see SI for more details on how we infer gender and geolocation information).

\subsection{Indicators of Sociopolitical Context}

To examine the socio-demographic determinants of abortion attitudes across the United States, we compile state- and county-level data from multiple sources. Demographic and economic indicators are drawn from the American Community Survey (ACS) \cite{acs_data} and include  percentage of residents with a bachelor's degree, median age, housing costs, income, and racial/ethnic composition of the population. Measures of abortion policy restrictiveness are sourced from the Guttmacher Institute \cite{guttmacher_2024}. We use county-level data on healthcare capacity, such as physicians per 100{,}000 residents—from the County Health Rankings and Roadmap \cite{countyhealth_2024}, while 2020 presidential vote share data is obtained from the MIT Election Lab \cite{mit_electionlab_2024}. Data on access to maternity care is drawn from the March of Dimes \cite{marchofdimes_2024}, religious congregation counts are from the U.S. Religious Census \cite{religious_census_2020}, and maternal mortality rates are obtained from the Centers for Disease Control and Prevention (CDC) \cite{cdc_maternal_2024}.

The selected indicators capture a cross-section of regional demographic, economic, and ideological conditions that shape reproductive attitudes and access. Economic context is represented by median income and median monthly housing costs, which offer insight into household stability and financial security. Social and family structure is reflected in the percentage of households with children, percentage divorced, percentage of the population living in rural areas, and median age—factors that help contextualize family norms, isolation, and life course dynamics that may shape reproductive preferences. To assess gendered autonomy and maternal health infrastructure, we include the maternal mortality rate (MMR), the number of physicians per 100,000 residents, access to maternity care, and the percentage of uninsured individuals—key indicators of healthcare access and reproductive risk. Educational attainment (percentage of the population with a bachelor’s degree) and racial/ethnic composition (percent Black, Latino, and Asian) further illuminate structural disparities that intersect with reproductive decision-making. Religiosity is measured by the proportion of the population affiliated with religious congregations, using data from the U.S. Religious Census. 

Finally, we capture policy and ideological climate through a binary indicator of abortion policy restrictiveness from the Guttmacher Institute and the \textit{Republican vote share} in the 2020 presidential election, obtained from the MIT Election Lab. Together, these variables provide a comprehensive view of offline regional conditions shaping online attitudes towards abortion in the United States.

\subsection{Inferring Gender}

We used a name-based gender classifier~\cite{van2023open} to infer user gender. This open-source classifier is based on Cultural Consensus Theory (CCT) and integrates 36 publicly available datasets spanning over 150 countries and more than a century of name-gender associations to produce probabilistic gender estimates from names. It employs an expectation-maximization algorithm to jointly infer the consensus gender association for each name and the reliability of each data source, without relying on ground-truth training labels. By weighting more reliable sources more heavily, the classifier generates robust gender predictions. 

We used the publicly available Python implementation \cite{nomquamgender} provided by the authors to classify gender from social media user names. We focus only on the two largest gender groups in the United States - men and women. This leaves us with a total of $35,949,368$ tweets from $2,877,409$ unique users. Additional descriptive statistics of tweet volumes, number of users and tweets per user disaggregated by gender are presented in Table \ref{tab:user_tweet_summary} under SI Section \ref{sec:additional}. Validation of the gender classifier is discussed under SI Section \ref{sec:gender_valid}. 

\subsection{Geolocation of Tweets}

We applied Carmen~\cite{dredze2013carmen}, a geo-location identification tool for Twitter data, to assign tweets to locations within the United States. Carmen leverages tweet metadata, including the ``place'' and ``coordinates'' fields, which provide information such as country, city, and geographic coordinates. It also incorporates location mentions from users' Twitter bios to improve inference. Manual validation confirmed the method's effectiveness in identifying users' home states. This process yielded a dataset of $9,982,073$ tweets from $805,382$ geo-located users across the U.S.

\subsection{Stance Classification}

Tweets with a pro-abortion stance endorse access to safe and legal abortion as a critical healthcare and human rights issue, while anti-abortion tweets frame it as a moral wrong that harms the unborn. Tweets with a neutral stance typically share news or legal updates about abortion access without expressing a clear stance. We fine-tune a LLaMA 3.1-8B Instruct model to classify abortion stance in tweets. The finetuning dataset combines 932 annotated tweets from the SemEval-2016 stance detection task \cite{mohammad2016semeval} and 1,000 additional tweets from our dataset manually labeled by two annotators as favor, against, or neutral/irrelevant~\cite{chang2023roeoverturned}. We use 75\% ($n=1418$) of the data for training and 25\% ($n=514$) for testing. Tweets are input using the prompt: \texttt{What is the stance expressed towards abortion right in the following tweet?}, with ground-truth labels from the annotations. The model achieves an overall accuracy of 78\% on the test set. Class-wise performance is as follows: for the favor class, precision = 0.79, recall = 0.78, and F1-score = 0.78 (support = 187); for neutral, precision = 0.81, recall = 0.79, and F1-score = 0.80 (support = 155); and for against, precision = 0.73, recall = 0.76, and F1-score = 0.74 (support = 172). The macro and weighted average F1-scores are both 0.78. Additional qualitative and quantitative validations are discussed in SI Section \ref{sec:stance_valid}.

\subsection{Emotion Detection}

We quantify emotions expressed in individual tweets using SpanEmo~\cite{alhuzali-ananiadou-2021-spanemo}, a transformer-based language model fine-tuned on the SemEval 2018 Task 1e-c dataset~\cite{mohammad-etal-2018-semeval}. SpanEmo detects six discrete emotions: \textit{anger}, \textit{disgust}, \textit{fear}, \textit{sadness}, \textit{joy}, and \textit{optimism}.  Validations are presented in SI Section \ref{sec:emo_valid}.

\subsection{User Ideology Classification}

To measure the ideological orientation of users independent of their abortion-related views, we analyze their activity in discussions surrounding the U.S. elections \cite{chen2022election2020}. We collect tweets from users who were active in the abortion debate and also participated in the broader political conversations between January 1, 2022 and January 1, 2023. This approach allows us to construct an ideology measure that is not confounded by abortion-specific attitudes. 

We identify $304,146$ users from the abortion dataset in \cite{chang2023roeoverturned} to be engaged in general U.S. political discourse during the same period \cite{chen2022election2020}. To infer ideological positions, we extract retweet interactions between these users and a curated set of 2,200 political elites whose ideological leanings are known (liberal = $-1$, conservative = $1$). Limiting the analysis to interactions occurring at least 10 times, we map these users into a latent ideological space following the approach in \cite{barbera2015birds, nettasinghe2025group} which is discussed in more depth in SI Section \ref{sec:ideo_valid}.

\subsection{Statistical Modeling}

We employ a series of linear and mixed-effects regression models to assess the demographic, ideological, temporal, and emotional dynamics associated with abortion-related discourse on social media. All models were estimated in Python using the \texttt{statsmodels} and \texttt{Pymer4} packages, with appropriate adjustments for random effects and interaction terms.

\paragraph{\textbf{Socio-demographic Correlates of the Gender Gap in Abortion Stance}}

To investigate the socio-demographic drivers of gender gaps in abortion stance, we model individual-level stance as a function of the  socio-demographic variables. The following mixed-effects model includes a random intercept for U.S. state to account for geographic clustering:

\begin{align*}
\texttt{Stance} &\sim \{\text{Education} + \text{Restrictiveness} + \text{Republican Voteshare}\\ &+ \text{Uninsured} 
+ \text{Households with Children} + \text{Divorced}\\ &+ \text{Rural} + \text{Cost of Housing} + \text{Median Income} \\ &+\text{Median Age} + \text{Black} + \text{Hispanic} + \text{Asian} \\ &+\text{Maternity Care Access} + \text{Religiosity} \\ &+ \text{Maternal Mortality Rate} \\
&+ \text{Physicians per 100K}\} \times \text{Gender} + (1 \mid \text{State})
\end{align*}

\paragraph{\textbf{Gender and Ideology as Predictors of Abortion Stance}}

We estimate another mixed-effects model to evaluate the relative effects of gender identity and user's ideological leaning on abortion stance:

\begin{align*}
\texttt{Stance} \sim \text{Ideology} + \text{Gender} + \text{Ideology} \times \text{Gender}+ (1 \mid \text{State})
\end{align*}

This model tests whether gender moderates the effect of ideological environment on individual stance, reinforcing the role of identity beyond partisan alignment.

\paragraph{\textbf{Emotional Tone by Gender-Stance Group and State Ideology}}

To examine the affective dimension of abortion discourse, we model the presence of various emotions (e.g., anger, disgust, fear, sadness, optimism, joy) using:

\begin{align*}
\texttt{Emotion} \sim \text{Group} + \text{State Lean} + \text{Group} \times \text{State Lean}
\end{align*}

Here, \texttt{Group} refers to combinations of gender and stance (e.g.,  Men and Women in Support of and Opposed to Abortion), and \texttt{State Lean} represents the partisan leaning of the user’s state of residence (Democratic/Republican) based on voteshare in the 2020 Presidential Elections.

\paragraph{\textbf{Temporal Dynamics of Pro-Abortion Stance}}

To evaluate changes in abortion stance before and after the leak of the \textit{Dobbs v. Jackson} decision, we model stance at the user level as a function of time (pre/post), gender, and state-level ideology:

\begin{align*}
\texttt{Stance} &\sim \text{Time} + \text{Gender} + \text{Ideology} \\
&+ \text{Time} \times \text{Gender} + \text{Time} \times \text{Ideology} \\
&+ \text{Gender} \times \text{Ideology} + \text{Time} \times \text{Gender} \times \text{Ideology}
\end{align*}

We distinguish between users active before the ruling and new users who joined the conversation post-\textit{Dobbs}, allowing us to assess whether observed stance shifts are driven by opinion change or new participation.

\paragraph{\textbf{Activity Levels by Gender-Stance Group Over Time}}

Finally, to compare activity levels across gender-stance groups, we model tweet volume as a function of group and time:

\begin{align*}
\texttt{Tweets} \sim \text{Group} + \text{Time} + \text{Group} \times \text{Time}
\end{align*}

This model compares the activity of each group  following the leak of the Supreme Court ruling, illuminating patterns of discursive mobilization.

\subsection*{Data Availability}

Due to Twitter's Terms of Service, we are unable to share tweet content directly. However, the tweet identifiers used in this study are publicly available at \url{https://dataverse.harvard.edu/dataset.xhtml?persistentId=doi:10.7910/DVN/STU0J5}.  The emotion classifier at \url{https://github.com/hasanhuz/SpanEmo}, the geolocation classifier at \url{https://github.com/mdredze/carmen}, and the gender classifier at \url{https://github.com/ianvanbuskirk/nomquamgender}. The stance and ideology classifiers developed for this study will be made publicly available upon acceptance of the manuscript. Additionally, we anonymized author usernames after classification and prior to conducting any downstream analyses to ensure privacy.

\clearpage 

\bibliography{reference,pnas-sample}

\begin{thebibliography}{10}
\providecommand{\url}[1]{\texttt{#1}}
\expandafter\ifx\csname urlstyle\endcsname\relax
  \providecommand{\doi}[1]{doi:\discretionary{}{}{}#1}\else
  \providecommand{\doi}{doi:\discretionary{}{}{}\begingroup \urlstyle{rm}\Url}\fi

\bibitem{rao2025polarized}
A.~Rao, R.-C. Chang, Q.~Zhong, K.~Lerman, M.~Wojcieszak, Polarized Online Discourse on Abortion: Frames and Hostile Expressions Among Liberals and Conservatives, in \emph{Proceedings of the International AAAI Conference on Web and Social Media}, vol.~19 (2025), pp. 1649--1668.

\bibitem{levendusky2024has}
M.~Levendusky, \emph{et~al.}, Has the Supreme Court become just another political branch? Public perceptions of court approval and legitimacy in a post-Dobbs world. \emph{Science Advances} \textbf{10}~(10), eadk9590 (2024).

\bibitem{gallup2010abortion}
L.~Saad, Republicans', Dems' Abortion Views Grow More Polarized, \url{https://news.gallup.com/poll/126374/republicans-dems-abortion-views-grow-polarized.aspx} (2010), [Online; accessed 11-January-2024].

\bibitem{gallup2011abortion}
L.~Saad, Americans Still Split Along Pro-Choice Pro-Life Lines, \url{https://news.gallup.com/poll/147734/americans-split-along-pro-choice-pro-life-lines.aspx} (2010), [Online; accessed 11-January-2024].

\bibitem{gallup2021abortion}
M.~Brenan, Record-High 47\% in U.S. Think Abortion Is Morally Acceptable, \url{https://news.gallup.com/poll/350756/record-high-think-abortion-morally-acceptable.aspx} (2010), [Online; accessed 11-January-2024].

\bibitem{dimaggio1996have}
P.~DiMaggio, J.~Evans, B.~Bryson, Have American's social attitudes become more polarized? \emph{American journal of Sociology} \textbf{102}~(3), 690--755 (1996).

\bibitem{evans2003have}
J.~H. Evans, Have Americans' attitudes become more polarized?—An update. \emph{Social Science Quarterly} \textbf{84}~(1), 71--90 (2003).

\bibitem{murray2024nbc}
M.~Murray, Final NBC News poll: Harris-Trump race neck and neck, with significant gender gap, \url{https://www.nbcnews.com/politics/2024-election/final-nbc-news-poll-harris-trump-race-neck-neck-significant-gender-gap-rcna178361} (2024), accessed May 5, 2025.

\bibitem{saad2024gallup}
L.~Saad, S.~E. Jones, S.~Fioroni, Exploring young women's leftward expansion, \url{https://news.gallup.com/poll/649826/exploring-young-women-leftward-expansion.aspx} (2024), accessed May 5, 2025.

\bibitem{loll2019differences}
D.~Loll, K.~S. Hall, Differences in abortion attitudes by policy context and between men and women in the World Values Survey. \emph{Women \& health} \textbf{59}~(5), 465--480 (2019).

\bibitem{gallup2023abortion}
Gallup, Abortion Trends by Gender (2023), \url{https://news.gallup.com/poll/245618/abortion-trends-gender.aspx}, accessed: 2025-03-14.

\bibitem{americansurveycenter2023abortion}
K.~Bowman, D.~Cox, Gender, Generation, and Abortion: Shifting Politics and Perspectives After Roe (2023), \url{https://www.americansurveycenter.org/research/gender-generation-and-abortion-shifting-politics-and-perspectives-after-roe/}, accessed: 2025-03-14.

\bibitem{lizotte2015abortion}
M.-K. Lizotte, The abortion attitudes paradox: Model specification and gender differences. \emph{Journal of Women, Politics \& Policy} \textbf{36}~(1), 22--42 (2015).

\bibitem{zhang2016gender}
A.~X. Zhang, S.~Counts, Gender and ideology in the spread of anti-abortion policy, in \emph{Proceedings of the 2016 CHI Conference on Human Factors in Computing Systems} (2016), pp. 3378--3389.

\bibitem{bode2017closing}
L.~Bode, Closing the gap: Gender parity in political engagement on social media. \emph{Information, communication \& society} \textbf{20}~(4), 587--603 (2017).

\bibitem{hu2023tweeting}
L.~Hu, M.~W. Kearney, C.~M. Frisby, Tweeting and retweeting: Gender discrepancies in discursive political engagement and influence on Twitter. \emph{Journal of Gender Studies} \textbf{32}~(5), 441--459 (2023).

\bibitem{philippe2024abortion}
O.~Philippe, \emph{et~al.}, Abortion and Miscarriage on Twitter: Sentiment and Polarity Analysis from a gendered perspective, in \emph{Proceedings of the 16th ACM Web Science Conference} (2024), pp. 311--319.

\bibitem{braveman2011social}
P.~Braveman, S.~Egerter, D.~R. Williams, The social determinants of health: coming of age. \emph{Annual review of public health} \textbf{32}~(1), 381--398 (2011).

\bibitem{dawes2020political}
D.~E. Dawes, \emph{The political determinants of health} (JHU Press) (2020).

\bibitem{jaidka2021information}
K.~Jaidka, J.~Eichstaedt, S.~Giorgi, H.~A. Schwartz, L.~H. Ungar, Information-seeking vs. sharing: Which explains regional health? An analysis of Google Search and Twitter trends. \emph{Telematics and Informatics} \textbf{59}, 101540 (2021).

\bibitem{aleksandric2024analyzing}
A.~Aleksandric, H.~I. Anderson, A.~Dangal, G.~M. Wilson, S.~Nilizadeh, Analyzing the Stance of Facebook Posts on Abortion Considering State-Level Health and Social Compositions, in \emph{Proceedings of the International AAAI Conference on Web and Social Media}, vol.~18 (2024), pp. 15--28.

\bibitem{chang2023roeoverturned}
R.-C. Chang, A.~Rao, Q.~Zhong, M.~Wojcieszak, K.~Lerman, \#RoeOverturned: Twitter Dataset on the Abortion Rights Controversy. \emph{Proceedings of the International AAAI Conference on Web and Social Media} \textbf{17}, 997--1005 (2023).

\bibitem{van2023open}
I.~Van~Buskirk, A.~Clauset, D.~B. Larremore, An open-source cultural consensus approach to name-based gender classification, in \emph{Proceedings of the International AAAI Conference on Web and Social Media}, vol.~17 (2023), pp. 866--877.

\bibitem{barbera2015birds}
P.~Barber{\'a}, Birds of the same feather tweet together: Bayesian ideal point estimation using Twitter data. \emph{Political analysis} \textbf{23}~(1), 76--91 (2015).

\bibitem{nettasinghe2025group}
B.~Nettasinghe, A.~Rao, B.~Jiang, A.~G. Percus, K.~Lerman, In-Group Love, Out-Group Hate: A Framework to Measure Affective Polarization via Contentious Online Discussions, in \emph{Proceedings of the ACM on Web Conference 2025} (2025), pp. 560--575.

\bibitem{alhuzali2021spanemo}
H.~Alhuzali, S.~Ananiadou, {S}pan{E}mo: Casting Multi-label Emotion Classification as Span-prediction. \emph{Proc. European Chapter of the ACL} pp. 1573--1584 (2021).

\bibitem{dredze2013carmen}
M.~Dredze, M.~J. Paul, S.~Bergsma, H.~Tran, Carmen: A twitter geolocation system with applications to public health. \emph{AAAI workshop on HIAI} \textbf{23}, 45 (2013).

\bibitem{rao2023pandemic}
A.~Rao, S.~Guo, S.~Y.~N. Wang, F.~Morstatter, K.~Lerman, Pandemic Culture Wars: Partisan Differences in the Moral Language of COVID-19 Discussions. \emph{2023 IEEE International Conference on Big Data (BigData)} pp. 413--422 (2023).

\bibitem{iyengar2015fear}
S.~Iyengar, S.~J. Westwood, Fear and loathing across party lines: New evidence on group polarization. \emph{American journal of political science} \textbf{59}~(3), 690--707 (2015).

\bibitem{nettasinghe2025out}
B.~Nettasinghe, A.~G. Percus, K.~Lerman, How out-group animosity can shape partisan divisions: A model of affective polarization. \emph{PNAS nexus} \textbf{4}~(3), pgaf082 (2025).

\bibitem{brady2021social}
W.~J. Brady, K.~McLoughlin, T.~N. Doan, M.~J. Crockett, How social learning amplifies moral outrage expression in online social networks. \emph{Science Advances} \textbf{7}~(33), eabe5641 (2021).

\bibitem{gerstein2022supreme}
J.~Gerstein, A.~Ward, Supreme Court has voted to overturn abortion rights, draft opinion shows (2022), \url{https://www.politico.com/news/2022/05/02/supreme-court-abortion-draft-opinion-00029473}, accessed: 2025-05-14.

\bibitem{pew2022partisan}
H.~Hartig, Wide partisan gaps in abortion attitudes, but opinions in both parties are complicated, \url{https://www.pewresearch.org/short-reads/2022/05/06/wide-partisan-gaps-in-abortion-attitudes-but-opinions-in-both-parties-are-complicated/} (2022), [Online; accessed 9-January-2024].

\bibitem{barbera2019leads}
P.~Barber{\'a}, \emph{et~al.}, Who leads? Who follows? Measuring issue attention and agenda setting by legislators and the mass public using social media data. \emph{American Political Science Review} \textbf{113}~(4), 883--901 (2019).

\bibitem{hanusch2019comments}
F.~Hanusch, E.~C. Tandoc~Jr, Comments, analytics, and social media: The impact of audience feedback on journalists’ market orientation. \emph{Journalism} \textbf{20}~(6), 695--713 (2019).

\bibitem{nelson2019doing}
J.~L. Nelson, E.~C. Tandoc~Jr, Doing “well” or doing “good”: What audience analytics reveal about journalism’s competing goals. \emph{Journalism Studies} \textbf{20}~(13), 1960--1976 (2019).

\bibitem{mcgregor2019social}
S.~C. McGregor, Social media as public opinion: How journalists use social media to represent public opinion. \emph{Journalism} \textbf{20}~(8), 1070--1086 (2019).

\bibitem{mcgregor2020taking}
S.~C. McGregor, “Taking the temperature of the room” how political campaigns use social media to understand and represent public opinion. \emph{Public Opinion Quarterly} \textbf{84}~(S1), 236--256 (2020).

\bibitem{acs_data}
{United States Census Bureau}, American Community Survey (ACS), \url{https://data.census.gov/} (2024), accessed June 2025.

\bibitem{guttmacher_2024}
{Guttmacher Institute}, State Abortion Policies, \url{https://states.guttmacher.org/policies/} (2024), accessed June 2025.

\bibitem{countyhealth_2024}
{University of Wisconsin Population Health Institute}, County Health Rankings and Roadmap, \url{https://www.countyhealthrankings.org/} (2024), accessed June 2025.

\bibitem{mit_electionlab_2024}
{MIT Election Data and Science Lab}, MIT Election Data and Science Lab: 2020 Presidential Vote Share, \url{https://electionlab.mit.edu/data} (2024), accessed June 2025.

\bibitem{marchofdimes_2024}
{March of Dimes}, Maternity Care Deserts Report, \url{https://www.marchofdimes.org/maternity-care-deserts-report} (2024), accessed June 2025.

\bibitem{religious_census_2020}
{Association of Statisticians of American Religious Bodies}, U.S. Religion Census: Religious Congregations and Membership Study, 2020, \url{https://www.usreligioncensus.org/maps2020_study} (2021), accessed June 2025.

\bibitem{cdc_maternal_2024}
{Centers for Disease Control and Prevention (CDC)}, Maternal Mortality Statistics, \url{https://www.cdc.gov/nchs/maternal-mortality/index.htm} (2024), accessed June 2025.

\bibitem{nomquamgender}
I.~V. Buskirk, Nomquamgender: An Open-Source Python Package for Name-Based Gender Classification (2022), \url{https://github.com/ianvanbuskirk/nomquamgender}, accessed: 2025-05-06.

\bibitem{mohammad2016semeval}
S.~Mohammad, S.~Kiritchenko, P.~Sobhani, X.~Zhu, C.~Cherry, Semeval-2016 task 6: Detecting stance in tweets, in \emph{Proceedings of the 10th international workshop on semantic evaluation (SemEval-2016)} (2016), pp. 31--41.

\bibitem{alhuzali-ananiadou-2021-spanemo}
H.~Alhuzali, S.~Ananiadou, {S}pan{E}mo: Casting Multi-label Emotion Classification as Span-prediction, in \emph{Proceedings of the 16th Conference of the European Chapter of the Association for Computational Linguistics: Main Volume}, P.~Merlo, J.~Tiedemann, R.~Tsarfaty, Eds. (Association for Computational Linguistics, Online) (2021), pp. 1573--1584, \doi{10.18653/v1/2021.eacl-main.135}, \url{https://aclanthology.org/2021.eacl-main.135}.

\bibitem{mohammad-etal-2018-semeval}
S.~Mohammad, F.~Bravo-Marquez, M.~Salameh, S.~Kiritchenko, {S}em{E}val-2018 Task 1: Affect in Tweets, in \emph{Proc. 12th Int. Workshop on Semantic Evaluation} (2018), pp. 1--17.

\bibitem{chen2022election2020}
E.~Chen, A.~Deb, E.~Ferrara, \# Election2020: The first public Twitter dataset on the 2020 US Presidential election. \emph{Journal of Computational Social Science} \textbf{5}~(1), 1--18 (2022).

\bibitem{prri2024abortionviews}
{Public Religion Research Institute (PRRI)}, Abortion Views in All 50 States: Findings from PRRI’s 2023 American Values Atlas, Online report with PDF slides (2024), \url{https://prri.org/research/abortion-views-in-all-50-states-findings-from-prris-2023-american-values-atlas/}, based on interviews with over 22,000 U.S. adults from March 9 to December 7, 2023.

\bibitem{plannedparenthood_scorecard}
{Planned Parenthood Action Fund}, Congressional Scorecard, \url{https://www.plannedparenthoodaction.org/congressional-scorecard} (2025), \url{https://www.plannedparenthoodaction.org/congressional-scorecard}, accessed: 2025-07-07.

\bibitem{chen2020tracking}
E.~Chen, K.~Lerman, E.~Ferrara, \emph{et~al.}, Tracking social media discourse about the covid-19 pandemic: Development of a public coronavirus twitter data set. \emph{JMIR public health and surveillance} \textbf{6}~(2), e19273 (2020).

\end{thebibliography}
\bibliographystyle{sciencemag}

\newpage

\section{Validations}
\label{sec:valid}

\subsection{Gender Classification}

To validate user gender, we leveraged self-disclosed gender identity markers in Twitter biographies. We  used regular expressions to detect explicit mentions of gender identity, including pronouns (e.g., she/her, he/him), gendered terms (e.g., woman, man). Specifically, we developed and applied a set of regular expression patterns to classify bios into two gender-relevant categories: woman and man. A user's biography was assigned a gender label if it matched any of the category-specific patterns. If no gender-relevant keywords were detected, the user's gender was left unassigned. This approach ensures that gender annotations reflect users’ self-identification. Table~\ref{tab:gender_terms} lists the terms used for gender identification. We treat these self-disclosed expressions as ground truth to evaluate the accuracy of the name-based gender classifier introduced in \cite{van2023open}  and discussed in the main text. The performance metrics are described in Table \ref{tab:gender_report}.

\begin{table}[ht]
\centering
\begin{tabular}{ll}
\hline
\textbf{Gender Category} & \textbf{Terms} \\
\hline
Woman & she/her, she/they, they/she, woman, female, girl, ladies, \\
       & lady, queen, mother, wife, sister, daughter, mom, aunt \\
\hline
Man   & he/him, he/they, they/he, man, male, boy, gentleman, \\
       & king, father, husband, brother, son , dad, uncle \\
\hline
\end{tabular}
\caption{Terms used to identify self-reported gender expressions in user biographies.}
\label{tab:gender_terms}
\end{table}

\begin{table}[!ht]
\centering
\begin{tabular}{lcccc}
\toprule
\textbf{Class} & \textbf{Precision} & \textbf{Recall} & \textbf{F1-score} & \textbf{Support} \\
\midrule
Man & 0.66 & 0.87 & 0.75 & 196,837 \\
Woman & 0.90 & 0.72 & 0.80 & 315,103 \\
\midrule
\textbf{Accuracy} &  &  & 0.78 & 511,940 \\
\textbf{Macro Avg} & 0.78 & 0.80 & 0.78 & 511,940 \\
\textbf{Weighted Avg} & 0.81 & 0.78 & 0.78 & 511,940 \\
\bottomrule
\end{tabular}
\caption{Performance metrics of a binary classification model predicting gender (Man vs. Woman). The table reports precision, recall, F1-score, and support for each class, along with overall accuracy, macro average, and weighted average scores.}
\label{tab:gender_report}
\end{table}
\label{sec:gender_valid}

\subsection{Abortion Stance Classification}

We used manual annotations of 1,000 randomly selected tweets by two expert annotators (Krippendorff’s alpha = 0.65 for the three-level classification: favor, against, neutral), alongside the gold-standard labeled data provided in \cite{mohammad2016semeval}, as ground truth for stance classification. Table \ref{tab:stance_tab} summarizes the classification metrics for the stance classification described in the Methods section. Table\ref{tab:stance_examples} illustrates the classification results by providing examples of tweets in support of abortion and opposed to abortion. 

\begin{table}[ht]
\centering
\begin{tabular}{lcccc}
\hline
\textbf{Class} & \textbf{Precision} & \textbf{Recall} & \textbf{F1-Score} & \textbf{Support} \\
\hline
Favor          & 0.79               & 0.78            & 0.78              & 187              \\
Neutral        & 0.81               & 0.79            & 0.80              & 155              \\
Against        & 0.73               & 0.76            & 0.74              & 172              \\
\hline
\textbf{Accuracy}     &               &                 & \textbf{0.78}     & 514              \\
\textbf{Macro Avg}    & 0.78          & 0.78            & 0.78              & 514              \\
\textbf{Weighted Avg} & 0.78          & 0.78            & 0.78              & 514              \\
\hline
\end{tabular}
\caption{Classification report showing precision, recall, F1-score, and support for abortion stance classification.}
\label{tab:stance_tab}
\end{table}

\begin{table}[!ht]
    \centering
    \begin{tabularx}{\textwidth}{X c}
        \toprule
        \textbf{Tweet} & \textbf{Classification} \\
        \midrule
        Forced birth in a country that refuses to protect children from being murdered at school.& Support\\
         With Roe v. Wade expected to end, women with high-risk pregnancies and their doctors are agonizing over what bar they would need to clear to legally justify an abortion in some states.  “How almost dead does someone need to be?” one physician asked.& Support\\
         In Alabama, the penalty for getting an abortion after you are raped is more severe than the penalty for raping someone.  How fucked up is that?& Support \\
         \midrule
         The only argument worth having on this issue, is that an abortion is an act of violence murdering an innocent child by choice.& Opposed \\
         Biden Democrats continue to disregard Americans' conscience rights to promote the left's radical transgender and abortion agenda.& Opposed \\
         I am pro-life because God is pro-life!   The news coming out of Washington today that Roe v Wade has been overturned means that hundreds of thousands of lives will be saved each year in the USA.& Opposed \\
    \bottomrule
    \end{tabularx}
    \caption{Examples of tweets classified by stance toward abortion.}
    \label{tab:stance_examples}
\end{table}

\begin{figure}[!ht]
    \centering
    \begin{subfigure}{0.48\linewidth}
        \centering
        \includegraphics[width=\linewidth]{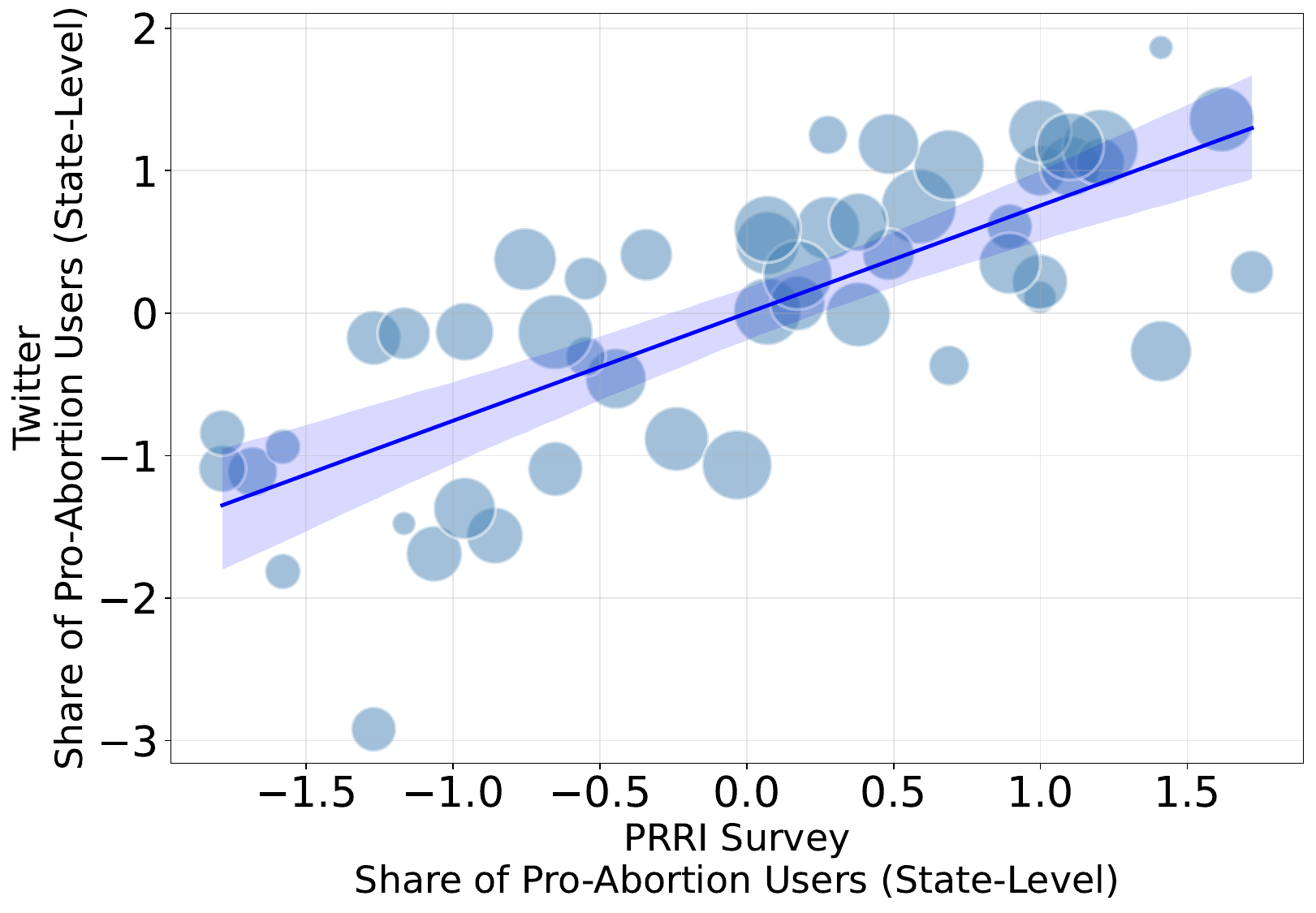}
        \caption{Comparing Twitter stance with Survey Responses at the state level.}
    \end{subfigure}
    \hfill
    \begin{subfigure}{0.48\linewidth}
        \centering
        \includegraphics[width=\linewidth]{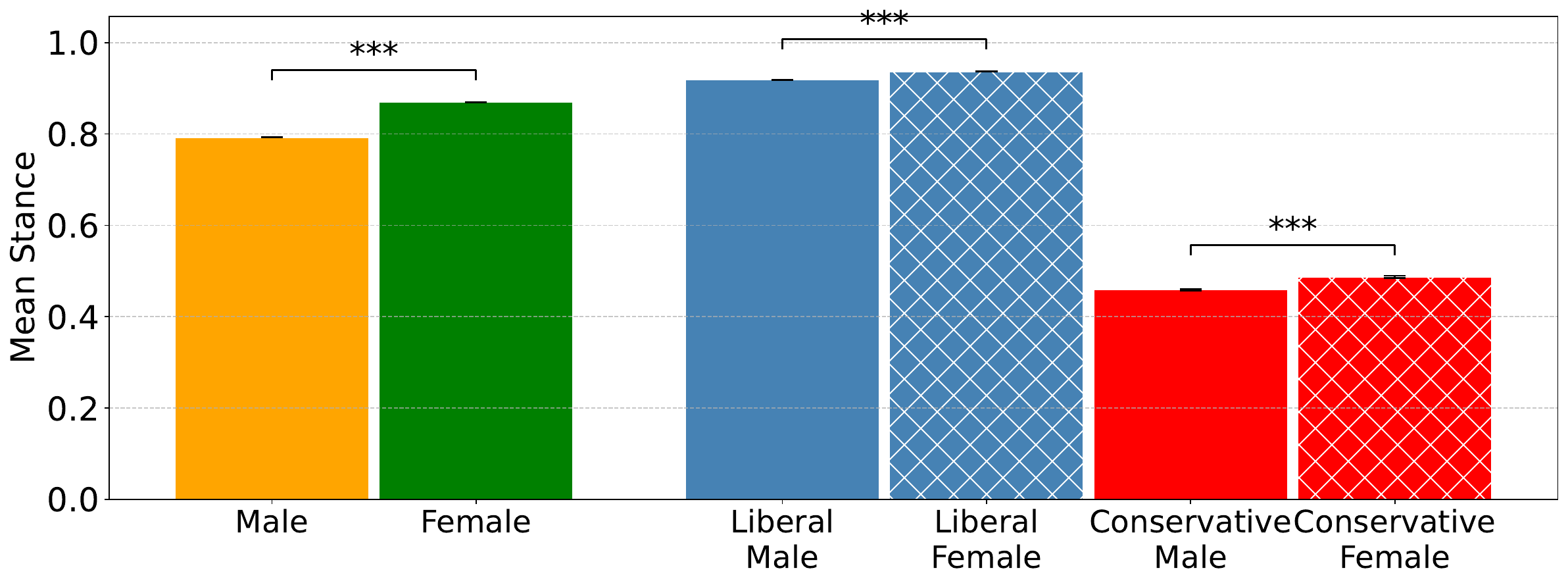}
        \caption{Comparing stance across groups.}
    \end{subfigure}
    \caption{(a) Correlation between state-level pro-abortion stance on Twitter and 2023 PRRI survey data, showing strong agreement (marker size reflects user volume).  (b) Twitter-based mean stance by gender and ideology, aligning with known trends: higher support among women and liberals, lower among conservatives. Significant differences marked (***, \(p < 0.001\)).
}
    \label{fig:stance_comparison}
\end{figure}

Figures \ref{fig:stance_comparison}(a) and  \ref{fig:stance_comparison}(b) demonstrate the validity of our stance classification model on Twitter using both aggregate survey data and individual-level group comparisons. In Figure \ref{fig:stance_comparison}(a), we compare our Twitter-derived estimates of pro-abortion stance at the state level to benchmark survey data from the 2023 Public Religion Research Institute (PRRI) American Values Atlas \cite{prri2024abortionviews}. The scatter plot shows a strong positive correlation (Pearson's r=0.76; regression slope=0.76) between the Z-normalized share of pro-abortion Twitter users and Z-normalized PRRI’s survey estimates across U.S. states. Each point represents a state, with marker size proportional to user volume. The significant linear relationship (blue regression line with 95\% CI) supports the external validity of Twitter-based inferences.

In Figure \ref{fig:stance_comparison}(b), we further assess construct validity by examining mean stance scores across gender and ideological user subgroups. Results align with well-established trends in public opinion research: women express more pro-abortion views than men, and liberals---particularly liberal women---express the strongest support ~\cite{loll2019differences,gallup2023abortion}. Conversely, conservative men and women show significantly lower mean stance scores. All pairwise differences marked with *** are statistically significant at $p < 0.001$.

Together, these validations provide robust evidence that our stance classification approach captures meaningful variation in public attitudes on abortion, consistent with both population-level survey data and subgroup-level ideological patterns.

In order to improve classification of abortion related tweets we extract the Congressional scorecard on abortion data from Planned Parenthood \cite{plannedparenthood_scorecard}. This provides the voting records of Senators and Representatives on abortion related bills. We then extract their tweets and assume to be in line with how they voted on average - support or against. This provides us with an augmented dataset of 86200 tweets (64667 in support of abortion, 20928 against abortion and 605 neutral). We use 70\% (n=60340) of this for supervised finetuning a LLaMA-3.1-8B-Instruct model and reserve 30\% (25860) for testing. We use this model to carry out our stance classifications.

\begin{table}[!hbt]
\centering
\begin{tabular}{lcccc}
\hline
\textbf{Class} & \textbf{Precision} & \textbf{Recall} & \textbf{F1-Score} & \textbf{Support} \\
\hline
Favor          & 0.93               & 0.95            & 0.94              & 19,322           \\
Neutral        & 0.79               & 0.79            & 0.79              & 155              \\
Against        & 0.84               & 0.78            & 0.81              & 6,383            \\
\hline
\textbf{Accuracy}     &               &                 & \textbf{0.91}     & 25,860           \\
\textbf{Macro Avg}    & 0.85          & 0.84            & 0.85              & 25,860           \\
\textbf{Weighted Avg} & 0.91          & 0.91            & 0.91              & 25,860           \\
\hline
\end{tabular}
\caption{Classification report with precision, recall, f1-score, and support for each class on augmented stance data.}
\label{tab:stance_aug}
\end{table}
\label{sec:stance_valid}

\subsection{Emotion Classification}

We validated emotions labels using manual coding. After two training sessions, two graduate students   independently coded the same 100 tweets to establish inter-coder reliability for anger (\(Cohen's \; \kappa = 0.631\)), disgust (\(Cohen's \; \kappa = 0.67\)), fear (\(Cronbach's \; \alpha = 0.63\)), sadness (\(Cohen's \; \kappa = 0.49\)), joy (\(Cohen's \; \kappa = 0.71\)), and optimism (\(Cohen's \; \kappa = 0.79\)). The two coders then coded 1000 tweets each to have a manual coding sample to validate the machine coding result. 

\begin{table}[!ht]
\centering
\scriptsize
\begin{tabular}{llcccc}
\toprule
\textbf{Emotion} & \textbf{Label} & \textbf{Precision} & \textbf{Recall} & \textbf{F1-Score} & \textbf{Support} \\
\midrule

\multirow{5}{*}{Anger} 
  & 0 (No Anger)   & 0.96 & 0.43 & 0.59 & 586 \\
  & 1 (Anger)      & 0.54 & 0.98 & 0.70 & 411 \\
  & Accuracy       &      &      & \textbf{0.65} & 997 \\
  & Macro Avg      & 0.75 & 0.70 & 0.65 & 997 \\
  & Weighted Avg   & 0.79 & 0.65 & 0.64 & 997 \\
\midrule

\multirow{5}{*}{Disgust} 
  & 0 (No Disgust) & 0.91 & 0.56 & 0.69 & 602 \\
  & 1 (Disgust)    & 0.60 & 0.93 & 0.73 & 395 \\
  & Accuracy       &      &      & \textbf{0.71} & 997 \\
  & Macro Avg      & 0.76 & 0.74 & 0.71 & 997 \\
  & Weighted Avg   & 0.79 & 0.71 & 0.71 & 997 \\
\midrule

\multirow{5}{*}{Fear} 
  & 0 (No Fear)    & 0.99 & 0.96 & 0.97 & 938 \\
  & 1 (Fear)       & 0.53 & 0.80 & 0.64 & 59 \\
  & Accuracy       &      &      & \textbf{0.95} & 997 \\
  & Macro Avg      & 0.76 & 0.88 & 0.81 & 997 \\
  & Weighted Avg   & 0.96 & 0.95 & 0.95 & 997 \\
\midrule

\multirow{5}{*}{Sadness} 
  & 0 (No Sadness) & 0.98 & 0.97 & 0.98 & 895 \\
  & 1 (Sadness)    & 0.77 & 0.83 & 0.80 & 102 \\
  & Accuracy       &      &      & \textbf{0.96} & 997 \\
  & Macro Avg      & 0.87 & 0.90 & 0.89 & 997 \\
  & Weighted Avg   & 0.96 & 0.96 & 0.96 & 997 \\
\midrule

\multirow{5}{*}{Joy} 
  & 0 (No Joy)     & 0.96 & 0.90 & 0.93 & 918 \\
  & 1 (Joy)        & 0.33 & 0.58 & 0.42 & 79 \\
  & Accuracy       &      &      & \textbf{0.87} & 997 \\
  & Macro Avg      & 0.65 & 0.74 & 0.67 & 997 \\
  & Weighted Avg   & 0.91 & 0.87 & 0.89 & 997 \\

\midrule
\multirow{5}{*}{Optimism} 
  & 0 (No Optimism) & 0.94 & 0.88 & 0.91 & 918 \\
  & 1 (Optimism)    & 0.37 & 0.63 & 0.46 & 79 \\
  & Accuracy        &      &      & \textbf{0.85} & 997 \\
  & Macro Avg       & 0.65 & 0.76 & 0.69 & 997 \\
  & Weighted Avg    & 0.90 & 0.85 & 0.87 & 997 \\

\bottomrule
\end{tabular}

\caption{Validating Emotions. Classification report with precision, recall, f1-score, and support for emotion.}
\label{tab:emotion_validation}
\end{table}

\label{sec:emo_valid}

\subsection{Ideology Classification}

In this model, retweets are treated as signals of agreement, positioning similar accounts closer together in the estimated space. The model accounts for elite popularity ($\lambda_j$), user political interest ($\eta_i$), and ideological alignment between elites and users ($\phi_j$, $\theta_i$), with the probability of a retweet expressed as:

$$
p(y_{i,j}=1 \mid \lambda_j,\eta_i,\gamma,\phi_j,\theta_i) = \text{logit}^{-1}(\lambda_j + \eta_i - \gamma(\phi_j - \theta_i)^2),
$$

where $\gamma$ serves as a regularization parameter. We estimate the model using Bayesian inference implemented in PyMC3, employing GPU acceleration and drawing 1,000 posterior samples across four chains following a 500-iteration warm-up. The resulting ideology scores for 32,000 users strongly agree with follower-based estimation in \cite{barbera2015birds} (Pearson $r=0.91$).

We obtain ideology estimates for 392,691 users who were active in the same period but discussing other topics related to U.S Politics in \cite{chen2022election2020}. Out of these, 307,130 were classified as liberal and 85,561 as conservative. 

\begin{figure}[ht]
    \centering
    \includegraphics[width=0.5\linewidth]{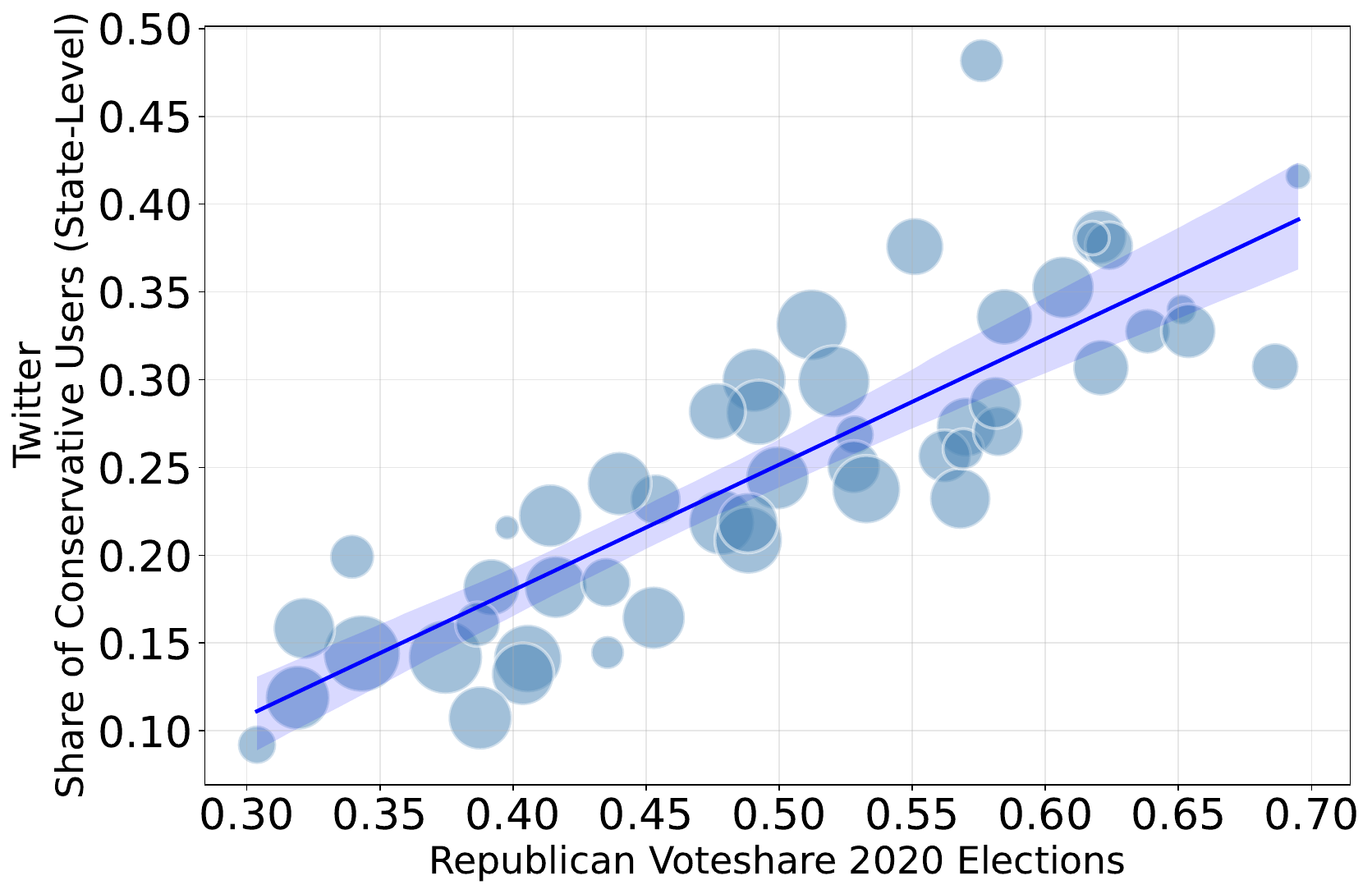}
    \caption{Correlation between state-level ideology estimation on Twitter and 2020 Republican voteshare in the Presidential Elections, showing strong agreement (marker size reflects user volume).}
    \label{fig:ideo_validation_state}
\end{figure}

Figures \ref{fig:ideo_validation_state} demonstrate the validity of the ideology classification model on Twitter using Republican vote-share from the 2020 U.S Presidential Elections. In Figure \ref{fig:ideo_validation_state}, we compare our Twitter-derived estimates of conservatism at the state level to against the 2020 Republican voteshare. The scatter plot shows a strong positive correlation (Pearson's r=0.86; regression slope=0.72) between the two across U.S. states. Each point represents a state, with marker size proportional to user volume. The significant linear relationship (blue regression line with 95\% CI) supports the external validity of Twitter-based inferences.

\label{sec:ideo_valid}

\begin{table}[!ht]
\centering
\scriptsize
\begin{tabular}{llcccc}
\toprule
\textbf{Metric} & \textbf{Group} & \textbf{All} & \textbf{Country} & \textbf{State} & \textbf{County} \\
\midrule
\multirow{3}{*}{Users} 
    & Men   & 1,603,351 & 452,623 & 421,675 & 267,009 \\
    & Women & 1,274,058 & 352,759 & 329,459 & 202,130 \\
    & Total  & 2,877,409 & 805,382 & 751,134 & 469,139 \\
\midrule
\multirow{3}{*}{Tweets} 
    & Men   & 18,735,034 & 5,185,865 & 4,697,225 & 2,763,759 \\
    & Women & 17,214,334 & 4,796,208 & 4,308,505 & 2,317,065\\
    & Total  & 35,949,368 & 9,982,073 & 9,005,730 & 5,080,824 \\
\midrule
\multirow{2}{*}{Tweets per User} 
    & Men   & 11.68 & 11.46 & 11.14 & 10.35 \\
    & Women & 13.51 & 13.60 & 13.08 & 11.46 \\
\midrule
\multirow{2}{*}{Abortion Stance: Users} 
    & Support & 2,223,178 & 642,200 & 605,030 & 389,385\\
    & Oppose  &   654,231 & 163,182 & 146,104 & 79,754 \\
\midrule
\multirow{2}{*}{Abortion Stance: Tweets} 
    & Support & 29,422,797 & 8,332,974 & 7,577,091 & 4,347,206 \\
    & Oppose  &  6,526,571 & 1,649,099 & 1,428,639 & 733,618 \\
\midrule
Tweets/User by Stance & Support & 13.23 & 12.98 & 12.52 & 11.16 \\
& Oppose & 9.98 & 10.11 & 9.78 & 9.20 \\
\midrule
\multirow{4}{*}{Users by Gender and Stance}
    & Men-Support   & 1,174,453 & 342,365 & 322,609 & 211,698 \\
    & Women-Support & 1,048,725 & 299,835 & 282,421 & 177,687 \\
    & Men-Opposed   &   428,898 & 110,258 & 99,066 & 55,311 \\
    & Women-Opposed &   225,333 &  52,924 &  47,038 &  24,443 \\
\midrule
\multirow{4}{*}{Tweets by Gender and Stance}
    & Men-Support   & 14,724,810 & 4,160,583 & 3,802,646 & 2,282,820 \\
    & Women-Support & 14,697,987 & 4,172,391 & 3,774,445& 2,064,386 \\
    & Men-Opposed   &  4,010,224 & 1,025,282 & 894,579 & 480,939 \\
    & Women-Opposed &  2,516,347 &  623,817 &  534,060 &  252,679 \\
\midrule
\multirow{4}{*}{Tweets/User by Gender \& Stance}
    & Men-Support   & 12.54 & 12.15 & 11.79 & 10.79 \\
    & Women-Support & 14.02 & 13.92 & 13.37 & 11.62 \\
    & Men-Opposed   & 9.35  & 9.30  & 9.03  & 8.70 \\
    & Women-Opposed & 11.17 & 11.79 & 11.35 & 10.34 \\
\bottomrule
\end{tabular}
\caption{User and Tweet Statistics by Geography and Gender}
\label{tab:user_tweet_summary}
\end{table}

\section{Additional Results}
\label{sec:additional}
\subsection{County-Level Results}
\label{sec:county}
\begin{figure}[!ht]
    \includegraphics[width=\linewidth]{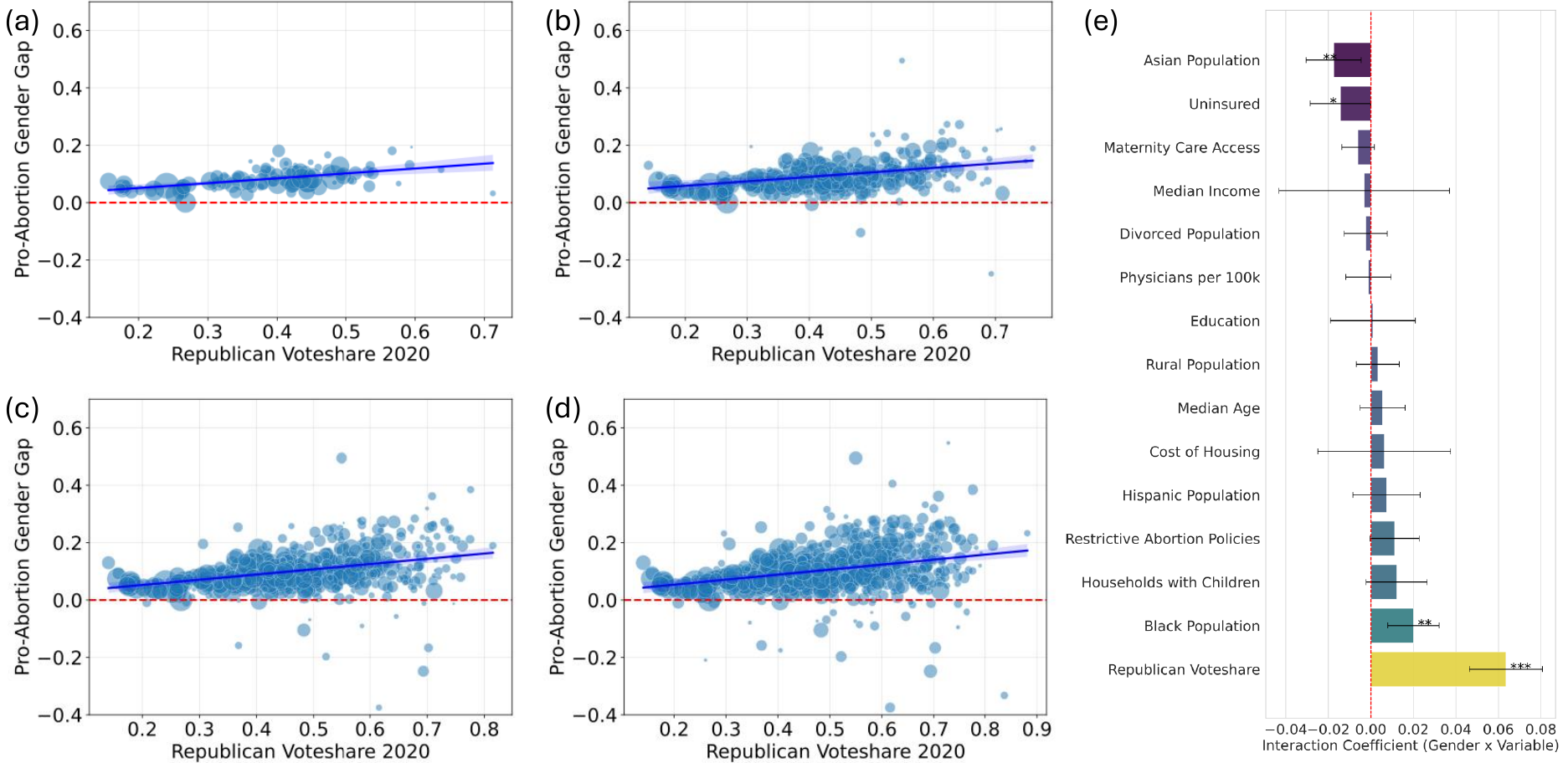}
    \caption{\textbf{Gender Gap in Support for Reproductive Rights at the County-Level.} (a-d) Gender gap in support for reproductive rights increases with Republican voteshare at the county-level for the top-100 (a), top-300 (b), top-500 (c) and top-700 (d) counties by population in the United States. 
    (e) Socio-demographic correlates of the gender gap in abortion stance. Republican vote share, percentage of Black and Asian and percentage of uninsured individuals are statistically significant predictors of the gender gap in abortion attitudes. Asterisks indicate statistical significance ($\ast p < 0.05$, $\ast\ast p < 0.01$, $\ast\ast\ast p < 0.001$).}
    \label{fig:gender_gap_county}
\end{figure}

While the main text focuses on state-level analyses, Figure~\ref{fig:gender_gap_county} extends the investigation to the county level to evaluate the robustness and granularity of the observed patterns. We examine the relationship between the gender gap in pro-abortion sentiment (ie., the difference in average support between women and men) and Republican vote share in the 2020 presidential election across the 100, 300, 500, and 700 most populous U.S. counties (Figures~\ref{fig:gender_gap_county}(a–d), respectively). Across all subsets, we find positive correlations between Republican support and the size of the gender gap, with stronger effects in more populous counties (Pearson’s $r = 0.56$, $0.43$, $0.41$, and $0.37$; all $p < 0.001$). These results parallel the state-level trend, suggesting that in more Republican-leaning counties, women are consistently more supportive of abortion rights than men. This pattern underscores the potential for gender identity to shape abortion attitudes independently of dominant partisan ideologies, even within politically conservative contexts.

We then explore the socio-demographic factors structuring these county-level gender gaps.  Figure~\ref{fig:gender_gap_county}(e) presentsinteraction model between gender and key county-level socio-demographic variables. Interaction coefficients assess whether a given socio-demographic variable is more strongly associated with pro-abortion attitudes among women than men. Several patterns align with the state-level findings. Higher percentage of uninsured residents are more negatively associated with women’s support relative to men’s. In counties with larger Black  populations, women exhibit significantly greater support than men, suggesting these environments may foster or enable the expression of more progressive reproductive attitudes among women than men. However, counties with higher Asian population show a reversed trend, where the difference between men and women reduces. Other variables do not show statistically significant gendered effects at the county level. Asterisks denote statistical significance $(\ast p < 0.05,\ \ast\ast p < 0.01,\ \ast\ast\ast p < 0.001)$.

\subsection{Changes Post Leak}
\label{sec:leak}
\begin{figure}[!ht]
    \centering
    \includegraphics[width=0.73\linewidth]{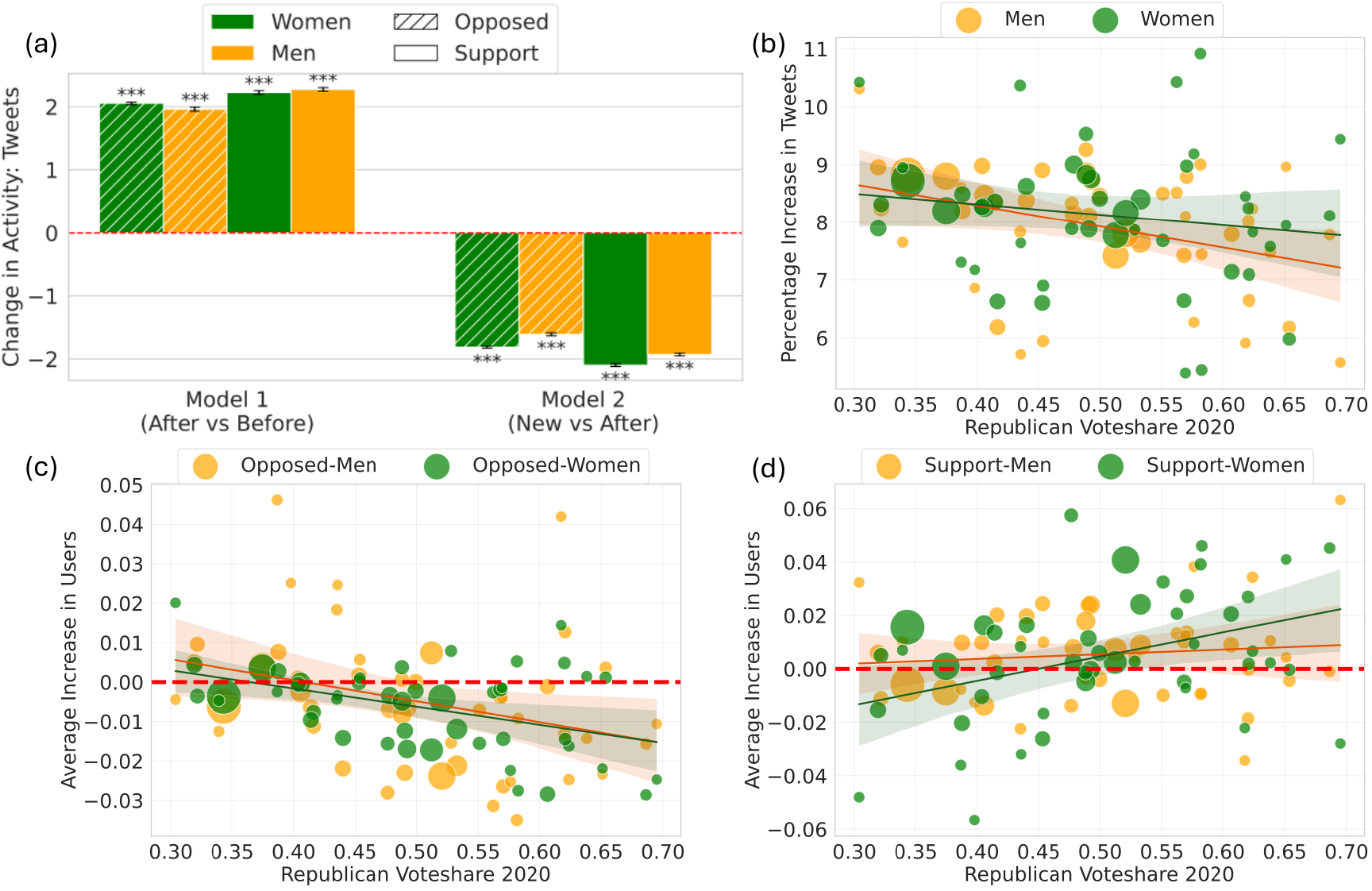}
    \caption{\textbf{Shifts in online engagement following the Supreme Court leak.}  (a) Among users active in both periods, all gender-stance groups increased tweet volume post-leak, with pro-abortion users showing the highest increases. New users who joined post-leak were far less active than pre-leak users. (b) Users in conservative states showed smaller activity increases, with minimal gender differences. (c) The share of anti-abortion users declined post-leak, especially in Republican-leaning states, with little gender variation. (d) The share of pro-abortion users increased post-leak, particularly in Republican-leaning states, again with minimal gender differences.}
    \label{fig:activity_change}
\end{figure}

To assess how the Dobbs leak changed user engagement, we model user tweet volume using negative binomial regression, stratified by gender and abortion stance (Men/Women $\times$ Support/Oppose Abortion), and interact these groupings with a binary time indicator (pre- vs. post-leak). Both within-subject and between-group analyses were conducted.

\begin{figure}[!ht]
    \centering
    \includegraphics[width=0.5\linewidth]{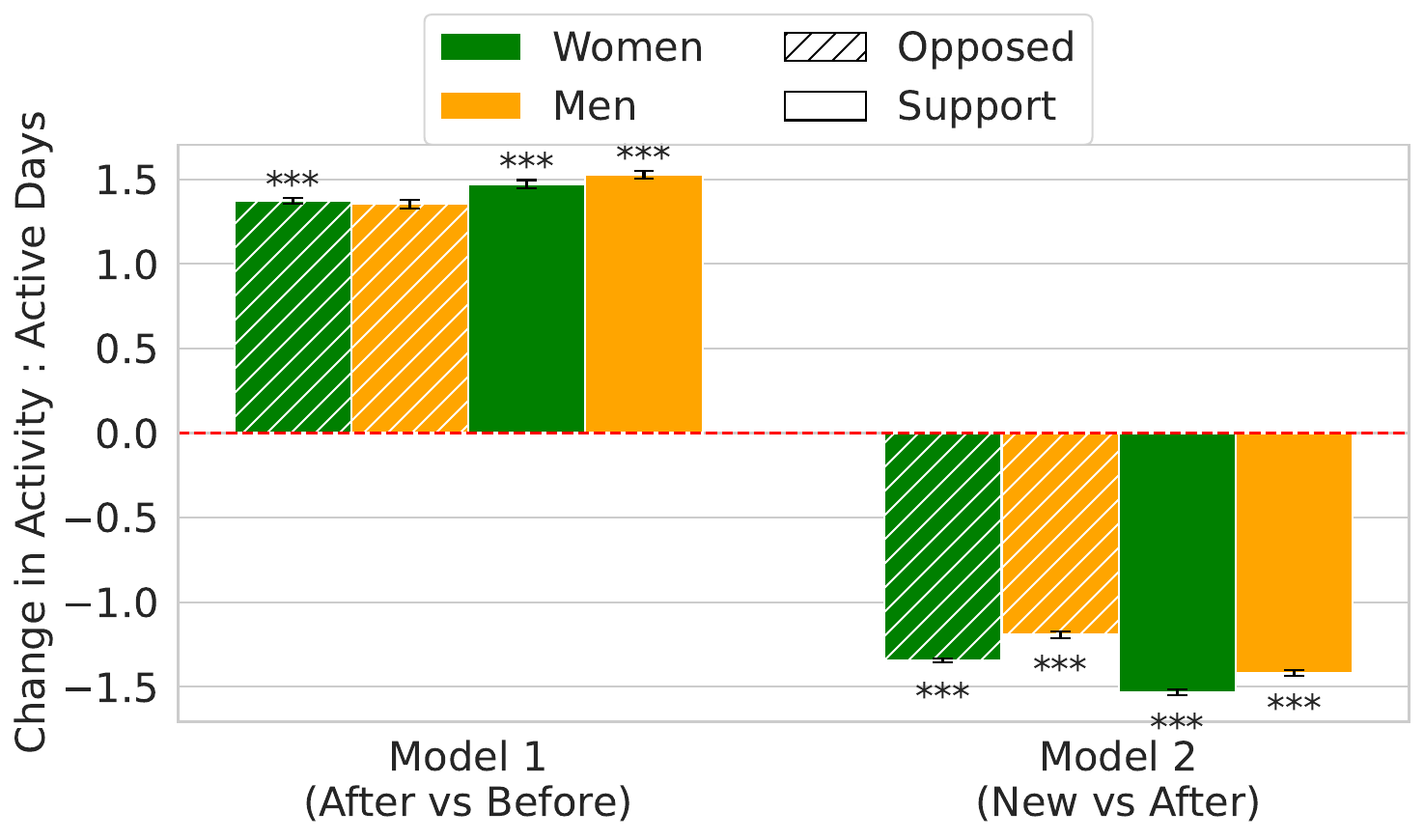}
    \caption{\textbf{Shifts in active days following the Supreme Court leak.}  (a) Among users active in both periods, all gender-stance groups increased in the number of days they were active post-leak, with pro-abortion users showing the highest increase. New users who joined post-leak were far less active than pre-leak users. (b) Users in conservative states showed smaller activity increases, with minimal gender differences. (c) The share of anti-abortion users declined post-leak, especially in Republican-leaning states, with little gender variation. (d) The share of pro-abortion users increased post-leak, particularly in Republican-leaning states, again with minimal gender differences.}
    \label{fig:days_active}
\end{figure}

As shown in Figure \ref{fig:activity_change}(a), all users who were active before the leak significantly increased their activity afterward, with pro-abortion users---both men and women---showing a larger increase than their anti-abortion counterparts. Supporters of abortion rights increased their tweet volume more than eightfold (from 4 to 33 tweets), compared to a sixfold increase among opponents. These patterns suggest that the leak disproportionately activated pro-abortion voices, especially among women.  Detailed regression results appear in Tables\ref{tab:nb_model_count_before_after} and\ref{tab:nb_model_count_before_after_new} in SI Section~\ref{sec:additional}. Comparing existing users and new participants, we find that newcomers were less active overall.  A follow-up analysis of the number of days users were active confirms these trends (Figure~\ref{fig:days_active}, SI Section~\ref{sec:additional}).

While men and women increased their activity by similar margins overall (Figure~\ref{fig:activity_change}(b)), users in more conservative states showed smaller gains—indicating that the response was somewhat muted in Republican-leaning regions.

Besides changes in activity patterns, the composition of online conversations also shifted after the leak. Figures~\ref{fig:activity_change}(c) and (d) show that in conservative states, newly mobilized users were more likely to support abortion rights and less likely to oppose them, relative to the already active user base. This selective mobilization further contributed to the increase in the gender gap, as newly engaged women—particularly in conservative environments—were disproportionately pro-abortion.

Together, these findings show that the Dobbs leak both intensified participation among existing users, especially those already supportive of abortion rights, and mobilized a new cohort of pro-abortion participants. The convergence of these trends---higher activity among supporters and selective engagement of pro-choice newcomers---increased the gender gap in abortion discourse, particularly in states where access to abortion was most under threat.

\begin{longtable}{lcccccc}

\toprule
\multicolumn{7}{c}{
\shortstack{
\texttt{Stance} $\sim$ (Education + Restrictiveness + Republican Voteshare + Uninsured + \\
Households with Children + Divorced + Rural Population + Cost of Housing + \\
Median Income + Median Age + Black Population + Hispanic Population + Asian Population + \\
Maternity Care Access + Religiosity + Maternal Mortality Rate + Physicians per 100K * Gender \\
+ (1 $\mid$ State)
}
} \\
\midrule
Intercept & -0.104 & 0.013 & -8.301 & 0.000 & -0.129 & -0.080 \\
Education & -0.023 & 0.020 & -1.136 & 0.256 & -0.061 & 0.016 \\
Restrictive Abortion Policies & -0.006 & 0.019 & -0.311 & 0.756 & -0.043 & 0.031 \\
Republican Voteshare & -0.138 & 0.031 & -4.432 & 0.000 & -0.200 & -0.077 \\
Uninsured & -0.008 & 0.025 & -0.320 & 0.749 & -0.056 & 0.040 \\
Households with Children & 0.051 & 0.021 & 2.467 & 0.014 & 0.011 & 0.092 \\
Divorced Population & 0.011 & 0.020 & 0.531 & 0.596 & -0.029 & 0.050 \\
Rural Population & -0.033 & 0.014 & -2.389 & 0.017 & -0.060 & -0.006 \\
Cost of Housing & -0.005 & 0.034 & -0.141 & 0.888 & -0.072 & 0.062 \\
Median Income & -0.021 & 0.029 & -0.716 & 0.474 & -0.078 & 0.036 \\
Median Age & 0.030 & 0.015 & 1.974 & 0.048 & 0.000 & 0.060 \\
Black Population & -0.047 & 0.019 & -2.539 & 0.011 & -0.084 & -0.011 \\
Hispanic Population & -0.050 & 0.028 & -1.781 & 0.075 & -0.104 & 0.005 \\
Asian Population & -0.018 & 0.013 & -1.426 & 0.154 & -0.043 & 0.007 \\
Maternity Care Access & -0.009 & 0.020 & -0.440 & 0.660 & -0.048 & 0.030 \\
Religiosity & 0.012 & 0.020 & 0.590 & 0.555 & -0.027 & 0.050 \\
Maternal Mortality Rate & -0.025 & 0.017 & -1.527 & 0.127 & -0.058 & 0.007 \\
Physicians per 100K & 0.015 & 0.015 & 1.007 & 0.314 & -0.014 & 0.044 \\
Gender & 0.240 & 0.002 & 102.146 & 0.000 & 0.235 & 0.245 \\
Education:Gender & 0.017 & 0.007 & 2.481 & 0.013 & 0.004 & 0.031 \\
Restrictive Abortion Policies:Gender & 0.008 & 0.007 & 1.196 & 0.232 & -0.005 & 0.021 \\
Republican Voteshare:Gender & 0.069 & 0.014 & 4.999 & 0.000 & 0.042 & 0.097 \\
Uninsured:Gender & -0.029 & 0.008 & -3.463 & 0.001 & -0.045 & -0.013 \\
Household with Children:Gender & -0.013 & 0.009 & -1.492 & 0.136 & -0.030 & 0.004 \\
Divorced Population:Gender & 0.010 & 0.007 & 1.591 & 0.112 & -0.002 & 0.023 \\
Rural Population:Gender & 0.021 & 0.006 & 3.462 & 0.001 & 0.009 & 0.032 \\
Cost of Housing:Gender & -0.027 & 0.009 & -2.886 & 0.004 & -0.046 & -0.009 \\
Median Income:Gender & 0.027 & 0.010 & 2.774 & 0.006 & 0.008 & 0.047 \\
Median Age:Gender & -0.013 & 0.006 & -2.265 & 0.024 & -0.025 & -0.002 \\
Black Population:Gender & 0.021 & 0.007 & 3.210 & 0.001 & 0.008 & 0.034 \\
Hispanic Population:Gender & 0.036 & 0.010 & 3.778 & 0.000 & 0.017 & 0.055 \\
Asian Population:Gender & 0.007 & 0.006 & 1.098 & 0.272 & -0.005 & 0.019 \\
Maternity Care Access:Gender & 0.009 & 0.007 & 1.269 & 0.204 & -0.005 & 0.023 \\
Religiosity:Gender & 0.001 & 0.008 & 0.164 & 0.869 & -0.014 & 0.016 \\
Maternal Mortality Rate:Gender & -0.001 & 0.005 & -0.111 & 0.911 & -0.009 & 0.008 \\
Physicians per 100K:Gender & -0.003 & 0.005 & -0.698 & 0.485 & -0.012 & 0.006 \\
\midrule
Group Var (state) & \multicolumn{6}{c}{0.003} \\
Residual Variance ($\sigma^2$) & \multicolumn{6}{c}{0.9713} \\
Number of Observations & \multicolumn{6}{c}{715,139} \\
Number of Groups (states) & \multicolumn{6}{c}{50} \\
\bottomrule
\caption{Mixed Effects Linear Regression of \texttt{Stance} with Demographics and Gender Interaction. State-level intercepts account for unobserved heterogeneity across states.}
\label{tab:mixedlm_socio}
\end{longtable}

\begin{table}[!ht]
\centering
\scriptsize
\begin{tabular}{lcccccc}
\toprule
\multicolumn{7}{c}{\texttt{stance} $\sim$ ideology + gender + ideology * gender + (1 $\mid$ state)} \\
\midrule
\textbf{Variable} & \textbf{Coef.} & \textbf{Std. Err.} & \textbf{z} & \textbf{P$>|$z$|$} & \textbf{[0.025} & \textbf{0.975]} \\
\midrule
Intercept             & -0.003 & 0.014 & -0.186 & 0.852 & -0.031 & 0.026 \\
\texttt{ideo}         & -0.610 & 0.001 & -465.430 & 0.000 & -0.613 & -0.608 \\
\texttt{gender}  & 0.034 & 0.001 & 27.187 & 0.000 & 0.032 & 0.037 \\
\texttt{ideology:gender} & 0.006 & 0.001 & 4.365 & 0.000 & 0.003 & 0.008 \\
\midrule
Group Var (state)     & \multicolumn{6}{c}{0.010} \\
Residual Variance ($\sigma^2$) & \multicolumn{6}{c}{0.6086} \\
Marginal $R^2$ (fixed effects only) & \multicolumn{6}{c}{0.441} \\
Conditional $R^2$ (fixed + random effects) & \multicolumn{6}{c}{0.456} \\
\bottomrule
\end{tabular}
\vspace{1em}
\begin{flushleft}
\caption{Mixed Effects Linear Regression predicting \texttt{stance} from ideology of users. This model specifies a random intercept model with fixed effects for ideology, gender, and their interaction. The intercept varies by U.S. state. Number of observations = 392,691; Number of groups = 51.}
\label{tab:mixedlm_ideo}
\end{flushleft}
\end{table}

\begin{table}[!ht]
\centering
\scriptsize
\begin{tabular}{llcccc}
\toprule
\textbf{Emotion} & \textbf{Group} & \textbf{All} & \textbf{Country} & \textbf{State} & \textbf{County} \\
\midrule
\multirow{5}{*}{Anger}
    & All Users       & 0.68 & 0.67 & 0.67 & 0.66 \\
    & Men-Support    & 0.68 & 0.67 & 0.67 & 0.66 \\
    & Women-Support  & 0.67 & 0.67 & 0.67 & 0.66 \\
    & Men-Opposed    & 0.72 & 0.70 & 0.70 & 0.69 \\
    & Women-Opposed  & 0.70 & 0.70 & 0.70 & 0.68 \\
\midrule
\multirow{5}{*}{Disgust}
    & All Users       & 0.68 & 0.67 & 0.67 & 0.66 \\
    & Men-Support    & 0.67 & 0.67 & 0.66 & 0.65 \\
    & Women-Support  & 0.67 & 0.66 & 0.66 & 0.65 \\
    & Men-Opposed    & 0.72 & 0.71 & 0.71 & 0.70 \\
    & Women-Opposed  & 0.71 & 0.70 & 0.70 & 0.69 \\
\midrule
\multirow{5}{*}{Fear}
    & All Users       & 0.14 & 0.14 & 0.14 & 0.14 \\
    & Men-Support    & 0.15 & 0.14 & 0.14 & 0.14 \\
    & Women-Support  & 0.16 & 0.15 & 0.15 & 0.15 \\
    & Men-Opposed    & 0.10 & 0.10 & 0.10 & 0.10 \\
    & Women-Opposed  & 0.11 & 0.12 & 0.11 & 0.11 \\
\midrule
\multirow{5}{*}{Sadness}
    & All Users       & 0.12 & 0.10 & 0.10 & 0.10\\
    & Men-Support    & 0.11 & 0.10 & 0.09 & 0.09 \\
    & Women-Support  & 0.12 & 0.10 & 0.10 & 0.10 \\
    & Men-Opposed    & 0.12 & 0.11 & 0.11 & 0.11 \\
    & Women-Opposed  & 0.14 & 0.13 & 0.13 & 0.12 \\
\midrule
\multirow{5}{*}{Joy}
    & All Users       & 0.08 & 0.08 & 0.08 & 0.08 \\
    & Men-Support    & 0.08 & 0.08 & 0.08 & 0.08 \\
    & Women-Support  & 0.08 & 0.08 & 0.09 & 0.09 \\
    & Men-Opposed    & 0.08 & 0.08 & 0.08 & 0.09 \\
    & Women-Opposed  & 0.09 & 0.08 & 0.08 & 0.09 \\
\midrule
\multirow{5}{*}{Optimism}
    & All Users       & 0.19 & 0.20 & 0.20 & 0.20 \\
    & Men-Support    & 0.19 & 0.20 & 0.20 & 0.20 \\
    & Women-Support  & 0.21 & 0.22 & 0.23 & 0.22 \\
    & Men-Opposed    & 0.14 & 0.14 & 0.14 & 0.15 \\
    & Women-Opposed  & 0.14 & 0.14 & 0.15 & 0.15 \\
\bottomrule
\end{tabular}
\caption{Emotion Percentages by Geography, Stance, and Gender}
\label{tab:emotion_percentages}
\end{table}

\scriptsize
\begin{longtable}[bht]{lrrrrrr}
\toprule
\textbf{Term} & \textbf{Coef.} & \textbf{Std.Err.} & \textbf{z} & \textbf{P$>$$|$z$|$} & \textbf{2.5\%} & \textbf{97.5\%}\\
\midrule
\multicolumn{7}{c}{Anger $\sim$ Group + Ideology + Group * Ideology} \\
\midrule
Intercept & -1.170 & 0.169 & -6.916 & 0.000 & -1.501 & -0.838 \\
C(Group)[Opposed\_Men] & -0.550 & 0.107 & -5.162 & 0.000 & -0.759 & -0.341 \\
C(Group)[Support\_Women] & 0.357 & 0.095 & 3.753 & 0.000 & 0.171 & 0.543 \\
C(Group)[Support\_Men] & 0.203 & 0.094 & 2.158 & 0.031 & 0.019 & 0.388 \\
Ideology & 0.050 & 0.036 & 1.406 & 0.160 & -0.020 & 0.120 \\
Ideology:C(Group)[Opposed\_Men] & -0.119 & 0.022 & -5.325 & 0.000 & -0.162 & -0.075 \\
Ideology:C(Group)[Support\_Women] & 0.082 & 0.020 & 4.142 & 0.000 & 0.043 & 0.121 \\
Ideology:C(Group)[Support\_Men] & 0.047 & 0.020 & 2.377 & 0.017 & 0.008 & 0.086 \\

\midrule

\multicolumn{7}{c}{Disgust $\sim$ Group + Ideology + Group * Ideology}\\
\midrule
Intercept & -1.216 & 0.169 & -7.210 & 0.000 & -1.546 & -0.885 \\
C(Group)[Opposed\_Men] & -0.585 & 0.106 & -5.511 & 0.000 & -0.794 & -0.377 \\
C(Group)[Support\_Women] & 0.333 & 0.095 & 3.515 & 0.000 & 0.148 & 0.519 \\
C(Group)[Support\_Men] & 0.209 & 0.094 & 2.222 & 0.026 & 0.025 & 0.393 \\
Ideology & 0.038 & 0.035 & 1.087 & 0.277 & -0.031 & 0.108 \\
Ideology:C(Group)[Opposed\_Men] & -0.126 & 0.022 & -5.651 & 0.000 & -0.169 & -0.082 \\
Ideology:C(Group)[Support\_Women] & 0.080 & 0.020 & 4.029 & 0.000 & 0.041 & 0.119 \\
Ideology:C(Group)[Support\_Men] & 0.050 & 0.020 & 2.555 & 0.011 & 0.012 & 0.089 \\
\midrule

\multicolumn{7}{c}{Fear $\sim$ Group + Ideology + Group * Ideology} \\
\midrule
Intercept & -0.467 & 0.125 & -3.728 & 0.000 & -0.713 & -0.222 \\
C(Group)[Opposed\_Men] & 0.062 & 0.079 & 0.789 & 0.430 & -0.092 & 0.217 \\
C(Group)[Support\_Women] & 0.194 & 0.070 & 2.751 & 0.006 & 0.056 & 0.332 \\
C(Group)[Support\_Men] & 0.036 & 0.070 & 0.515 & 0.607 & -0.101 & 0.173 \\
Ideology & 0.006 & 0.026 & 0.229 & 0.819 & -0.046 & 0.058 \\
Ideology:C(Group)[Opposed\_Men] & 0.015 & 0.017 & 0.904 & 0.366 & -0.017 & 0.047 \\
Ideology:C(Group)[Support\_Women] & 0.025 & 0.015 & 1.712 & 0.087 & -0.004 & 0.054 \\
Ideology:C(Group)[Support\_Men] & -0.003 & 0.015 & -0.182 & 0.856 & -0.031 & 0.026 \\
\midrule

\multicolumn{7}{c}{Sadness $\sim$ Group + Ideology + Group * Ideology} \\
\midrule
Intercept & -0.470 & 0.128 & -3.659 & 0.000 & -0.722 & -0.218 \\
C(Group)[Opposed\_Men] & 0.070 & 0.081 & 0.860 & 0.390 & -0.089 & 0.228 \\
C(Group)[Support\_Women] & 0.337 & 0.072 & 4.663 & 0.000 & 0.195 & 0.478 \\
C(Group)[Support\_Men] & 0.216 & 0.072 & 3.025 & 0.002 & 0.076 & 0.357 \\
Ideology & -0.012 & 0.027 & -0.432 & 0.666 & -0.065 & 0.041 \\
Ideology:C(Group)[Opposed\_Men] & 0.019 & 0.017 & 1.099 & 0.272 & -0.015 & 0.052 \\
Ideology:C(Group)[Support\_Women] & 0.069 & 0.015 & 4.580 & 0.000 & 0.040 & 0.099 \\
Ideology:C(Group)[Support\_Men] & 0.050 & 0.015 & 3.330 & 0.001 & 0.020 & 0.079 \\
\midrule

\multicolumn{7}{c}{Optimism $\sim$ Group + Ideology + Group * Ideology}\\
\midrule
Intercept & -0.350 & 0.144 & -2.430 & 0.015 & -0.633 & -0.068 \\
C(Group)[Opposed\_Men] & 0.367 & 0.091 & 4.045 & 0.000 & 0.189 & 0.545 \\
C(Group)[Support\_Women] & -0.590 & 0.081 & -7.287 & 0.000 & -0.749 & -0.431 \\
C(Group)[Support\_Men] & -0.535 & 0.080 & -6.669 & 0.000 & -0.693 & -0.378 \\
Ideology & 0.034 & 0.030 & 1.115 & 0.265 & -0.026 & 0.093 \\
Ideology:C(Group)[Opposed\_Men] & 0.078 & 0.019 & 4.134 & 0.000 & 0.041 & 0.116 \\
Ideology:C(Group)[Support\_Women] & -0.132 & 0.017 & -7.785 & 0.000 & -0.165 & -0.099 \\
Ideology:C(Group)[Support\_Men] & -0.117 & 0.017 & -6.941 & 0.000 & -0.149 & -0.084 \\
\midrule

\multicolumn{7}{c}{Joy $\sim$ Group + Ideology + Group * Ideology} \\
\midrule
Intercept & -0.215 & 0.098 & -2.194 & 0.028 & -0.407 & -0.023 \\
C(Group)[Opposed\_Men] & 0.250 & 0.062 & 4.060 & 0.000 & 0.130 & 0.371 \\
C(Group)[Support\_Women] & -0.321 & 0.055 & -5.831 & 0.000 & -0.429 & -0.213 \\
C(Group)[Support\_Men] & -0.264 & 0.055 & -4.833 & 0.000 & -0.371 & -0.157 \\
Ideology & 0.021 & 0.021 & 1.029 & 0.303 & -0.019 & 0.062 \\
Ideology:C(Group)[Opposed\_Men] & 0.053 & 0.013 & 4.129 & 0.000 & 0.028 & 0.079 \\
Ideology:C(Group)[Support\_Women] & -0.063 & 0.012 & -5.434 & 0.000 & -0.085 & -0.040 \\
Ideology:C(Group)[Support\_Men] & -0.050 & 0.011 & -4.344 & 0.000 & -0.072 & -0.027 \\
\bottomrule
\caption{Regression tables for assessing difference in emotions for the four groups of users - Men-Opposed, Men-Support, Women-Opposed and Women-Support} \\
\label{tab:emotion_regressions}
\end{longtable}

\begin{table}[!ht]
\centering
\scriptsize
\begin{tabular}{lcccccc}
\toprule
\multicolumn{7}{c}{\texttt{Stance} $\sim$ Time + Gender + Ideology + Time * Gender + Time*Ideology + Gender*Ideology +  Time * Gender * Ideology} \\
\midrule
\textbf{Variable} & \textbf{Coef.} & \textbf{Std. Err.} & \textbf{t} & \textbf{P$>|$t$|$} & \textbf{[0.025} & \textbf{0.975]} \\
\midrule
Intercept                       & 0.784 & 0.002 & 472.849 & 0.000 & 0.780 & 0.787 \\
\texttt{Time}             & 0.019 & 0.002 & 8.156   & 0.000 & 0.015 & 0.024 \\
\texttt{Gender}            & 0.061 & 0.002 & 24.510  & 0.000 & 0.056 & 0.065 \\
\texttt{Time:Gender} & 0.002 & 0.003 & 0.491   & 0.624 & -0.005 & 0.009 \\
\texttt{Ideology}                    & -0.071 & 0.003 & -25.426 & 0.000 & -0.076 & -0.066 \\
\texttt{Time:Ideology}         & 0.005 & 0.004 & 1.214   & 0.225 & -0.003 & 0.013 \\
\texttt{Gender:Ideology}        & 0.020 & 0.004 & 4.819   & 0.000 & 0.012 & 0.029 \\
\texttt{Time:Gender:Ideology} & -0.002 & 0.006 & -0.293 & 0.770 & -0.013 & 0.010 \\
\midrule
$R^2$ & \multicolumn{6}{c}{0.019} \\
Adjusted $R^2$ & \multicolumn{6}{c}{0.019} \\
\bottomrule
\end{tabular}
\caption{Ordinary Least Squares regression of user-level \texttt{Stance} on time period (Before vs. After Dobbs ruling), gender (1 = Women), and state-level ideology (1 = Republican-leaning, defined as >50\% voting Republican in the 2020 presidential election), including all two- and three-way interactions.}
\label{tab:stance_before_after}
\end{table}

\begin{table}[!ht]
\centering
\scriptsize
\begin{tabular}{lcccccc}
\toprule
\multicolumn{7}{c}{\texttt{Stance} $\sim$ Time + Gender + Ideology + Time * Gender + Time*Ideology + Gender*Ideology +  Time * Gender * Ideology}\\
\midrule
\textbf{Variable} & \textbf{Coef.} & \textbf{Std. Err.} & \textbf{t} & \textbf{P$>|$t$|$} & \textbf{[0.025} & \textbf{0.975]} \\
\midrule
Intercept                       & 0.803 & 0.002 & 524.449 & 0.000 & 0.800 & 0.806 \\
\texttt{Time}             & 0.019 & 0.002 & 11.147  & 0.000 & 0.016 & 0.022 \\
\texttt{Gender}            & 0.062 & 0.002 & 27.289  & 0.000 & 0.058 & 0.067 \\
\texttt{Time:Gender} & 0.003 & 0.003 & 1.144   & 0.253 & -0.002 & 0.008 \\
\texttt{Ideology}                    & -0.066 & 0.003 & -25.671 & 0.000 & -0.071 & -0.061 \\
\texttt{Time:Ideology}         & -0.001 & 0.003 & -0.226 & 0.821 & -0.006 & 0.005 \\
\texttt{Gender:Ideology}        & 0.019 & 0.004 & 4.769   & 0.000 & 0.011 & 0.026 \\
\texttt{Time:Gender:Ideology} & 0.014 & 0.004 & 3.267 & 0.001 & 0.006 & 0.023 \\
\midrule
$R^2$ & \multicolumn{6}{c}{0.022} \\
Adjusted $R^2$ & \multicolumn{6}{c}{0.022} \\
\bottomrule
\end{tabular}
\caption{Ordinary Least Squares regression of user-level \texttt{Stance} on time period (Old Users vs. New Users After Dobbs ruling), gender (1 = Women), and state-level ideology (1 = Republican-leaning, defined as >50\% voting Republican in the 2020 presidential election), including all two- and three-way interactions.}
\label{tab:stance_before_after_new}
\end{table}

\begin{table}[ht]
\centering
\scriptsize
\begin{tabular}{lcccccc}
\toprule
\multicolumn{7}{c}{\texttt{Tweets} $\sim$ Group + Time + Group * Time} \\
\midrule
\textbf{Variable} & \textbf{Coef.} & \textbf{Std. Err.} & \textbf{z} & \textbf{P$>|$z$|$} & \textbf{[0.025} & \textbf{0.975]} \\
\midrule
Intercept                         & 1.394 & 0.014 & 101.122 & 0.000 & 1.367 & 1.421 \\
\texttt{Group[T.Opposed\_Men]}   & -0.057 & 0.017 & -3.285 & 0.001 & -0.090 & -0.023 \\
\texttt{Group[T.Support\_Women]} & 0.251 & 0.015 & 16.542 & 0.000 & 0.221 & 0.280 \\
\texttt{Group[T.Support\_Men]}   & 0.027 & 0.015 & 1.806 & 0.071 & -0.002 & 0.057 \\
\texttt{Time}                     & 2.053 & 0.020 & 102.019 & 0.000 & 2.013 & 2.092 \\
\texttt{Opposed\_Men:Time}       & -0.091 & 0.025 & -3.659 & 0.000 & -0.140 & -0.042 \\
\texttt{Support\_Women:Time}     & 0.173 & 0.022 & 7.908 & 0.000 & 0.130 & 0.216 \\
\texttt{Support\_Men:Time}       & 0.221 & 0.022 & 10.213 & 0.000 & 0.179 & 0.264 \\
\midrule
$\alpha$ (dispersion) & \multicolumn{6}{c}{1.423 (SE = 0.004)} \\
Pseudo $R^2$ & \multicolumn{6}{c}{0.073} \\
Observations & \multicolumn{6}{c}{224,838} \\
\bottomrule
\end{tabular}
\caption{Negative Binomial regression of tweet count on group (opposed male, opposed female, support male, support female) and time (pre-leak vs post-leak).}
\label{tab:nb_model_count_before_after}
\end{table}

\begin{table}[ht]
\centering
\scriptsize
\begin{tabular}{lcccccc}
\toprule
\multicolumn{7}{c}{\texttt{Tweets} $\sim$ Group + Time + Group * Time} \\
\midrule
\textbf{Variable} & \textbf{Coef.} & \textbf{Std. Err.} & \textbf{z} & \textbf{P$>|$z$|$} & \textbf{[0.025} & \textbf{0.975]} \\
\midrule
Intercept                         & 3.447 & 0.013 & 262.561 & 0.000 & 3.421 & 3.473 \\
\texttt{Group[T.Opposed\_Men]}   & -0.148 & 0.016 & -9.190 & 0.000 & -0.179 & -0.116 \\
\texttt{Group[T.Support\_Women]} & 0.423 & 0.014 & 30.002 & 0.000 & 0.396 & 0.451 \\
\texttt{Group[T.Support\_Men]}   & 0.249 & 0.014 & 17.751 & 0.000 & 0.221 & 0.276 \\
\texttt{Time}                     & -1.813 & 0.015 & -121.948 & 0.000 & -1.842 & -1.783 \\
\texttt{Opposed\_Men:Time}       & 0.203 & 0.018 & 11.197 & 0.000 & 0.168 & 0.239 \\
\texttt{Support\_Women:Time}     & -0.285 & 0.016 & -17.863 & 0.000 & -0.316 & -0.254 \\
\texttt{Support\_Men:Time}       & -0.117 & 0.016 & -7.364 & 0.000 & -0.148 & -0.086 \\
\midrule
$\alpha$ (dispersion) & \multicolumn{6}{c}{1.135 (SE = 0.002)} \\
Pseudo $R^2$ & \multicolumn{6}{c}{0.076} \\
Observations & \multicolumn{6}{c}{619,960} \\
\bottomrule
\end{tabular}
\caption{Negative Binomial regression of tweet count on group (opposed male, opposed female, support male, support female) and time (old users vs new users).}
\label{tab:nb_model_count_before_after_new}
\end{table}

\begin{table}[!ht]
\centering
\scriptsize
\begin{tabular}{lcccccc}
\toprule
\multicolumn{7}{c}{\texttt{Days Active} $\sim$ Group + Time + Group * Time} \\
\midrule
\textbf{Variable} & \textbf{Coef.} & \textbf{Std. Err.} & \textbf{z} & \textbf{P$>|$z$|$} & \textbf{[0.025} & \textbf{0.975]} \\
\midrule
Intercept                          & 0.958 & 0.011 & 83.599 & 0.000 & 0.936 & 0.981 \\
\texttt{group[T.Opposed\_Men]}    & -0.064 & 0.014 & -4.468 & 0.000 & -0.092 & -0.036 \\
\texttt{group[T.Support\_Women]}  & 0.193 & 0.013 & 15.352 & 0.000 & 0.168 & 0.218 \\
\texttt{group[T.Support\_Men]}    & 0.047 & 0.012 & 3.726 & 0.000 & 0.022 & 0.071 \\
\texttt{Time}                      & 1.373 & 0.016 & 85.150 & 0.000 & 1.341 & 1.404 \\
\texttt{Opposed\_Men:Time}        & -0.018 & 0.020 & -0.924 & 0.356 & -0.058 & 0.021 \\
\texttt{Support\_Women:Time}      & 0.098 & 0.017 & 5.615  & 0.000 & 0.064 & 0.132 \\
\texttt{Support\_Men:Time}        & 0.153 & 0.017 & 8.797  & 0.000 & 0.119 & 0.187 \\
\midrule
$\alpha$ (dispersion) & \multicolumn{6}{c}{0.772 (SE = 0.003)} \\
Pseudo $R^2$ & \multicolumn{6}{c}{0.073} \\
Observations & \multicolumn{6}{c}{224,838} \\
\bottomrule
\end{tabular}
\caption{Negative Binomial regression of number of active days on Twitter as a function of  group (opposed male, opposed female, support male, support female) and time (pre-leak users vs post-leak users).}
\label{tab:nb_model}
\end{table}

\begin{table}[!ht]
\centering
\scriptsize
\begin{tabular}{lcccccc}
\toprule
\multicolumn{7}{c}{\texttt{days\_active} $\sim$ Group + Time + Group * Time} \\
\midrule
\textbf{Variable} & \textbf{Coef.} & \textbf{Std. Err.} & \textbf{z} & \textbf{P$>|$z$|$} & \textbf{[0.025} & \textbf{0.975]} \\
\midrule
Intercept                         & 2.331 & 0.010 & 236.589 & 0.000 & 2.312 & 2.350 \\
\texttt{Group[T.Opposed\_Men]}   & -0.083 & 0.012 & -6.845 & 0.000 & -0.106 & -0.059 \\
\texttt{Group[T.Support\_Women]} & 0.291 & 0.011 & 27.541 & 0.000 & 0.270 & 0.312 \\
\texttt{Group[T.Support\_Men]}   & 0.199 & 0.010 & 18.986 & 0.000 & 0.179 & 0.220 \\
\texttt{Time}                     & -1.343 & 0.011 & -117.265 & 0.000 & -1.365 & -1.320 \\
\texttt{Opposed\_Men:Time}       & 0.151 & 0.014 & 10.820 & 0.000 & 0.124 & 0.179 \\
\texttt{Support\_Women:Time}     & -0.190 & 0.012 & -15.494 & 0.000 & -0.214 & -0.166 \\
\texttt{Support\_Men:Time}       & -0.074 & 0.012 & -6.062 & 0.000 & -0.098 & -0.050 \\
\midrule
$\alpha$ (dispersion) & \multicolumn{6}{c}{0.560 (SE = 0.001)} \\
Pseudo $R^2$ & \multicolumn{6}{c}{0.081} \\
Observations & \multicolumn{6}{c}{619,960} \\
\bottomrule
\end{tabular}
\caption{Negative Binomial regression of number of active days on Twitter as a function of  group (opposed male, opposed female, support male, support female) and time (old users vs new users).}
\label{tab:nb_model_large}
\end{table}

\subsection{Gender Gap in Pro-Masking Attitudes}
\label{sec:masking}

We draw on publicly available Twitter data related to the COVID-19 pandemic, spanning from January 21, 2020 to November 4, 2021~\cite{chen2020tracking}. Following the approach in~\cite{nettasinghe2025group,rao2023pandemic}, we limit our analysis to tweets posted between January 21, 2020 and January 1, 2021, resulting in a corpus of approximately 230 million tweets from 8.7 million U.S.-based users. We focus on masking, an issue that had grown contentious during the pandemic and includes discussions about face coverings, mask mandates, mask shortages, and anti-mask sentiment. To identify tweets relevant to these topics, we use a weakly supervised keyword mining technique that extracts candidate terms from Wikipedia pages associated with each sub-issue~\cite{rao2023pandemic}. After manual validation of the extracted terms, we filter the dataset to retain tweets containing these keywords, yielding a final subset of 7.9 million tweets from 1.9 million geolocated users.

\begin{figure}[!ht]
    \centering
    \includegraphics[width=0.8\linewidth]{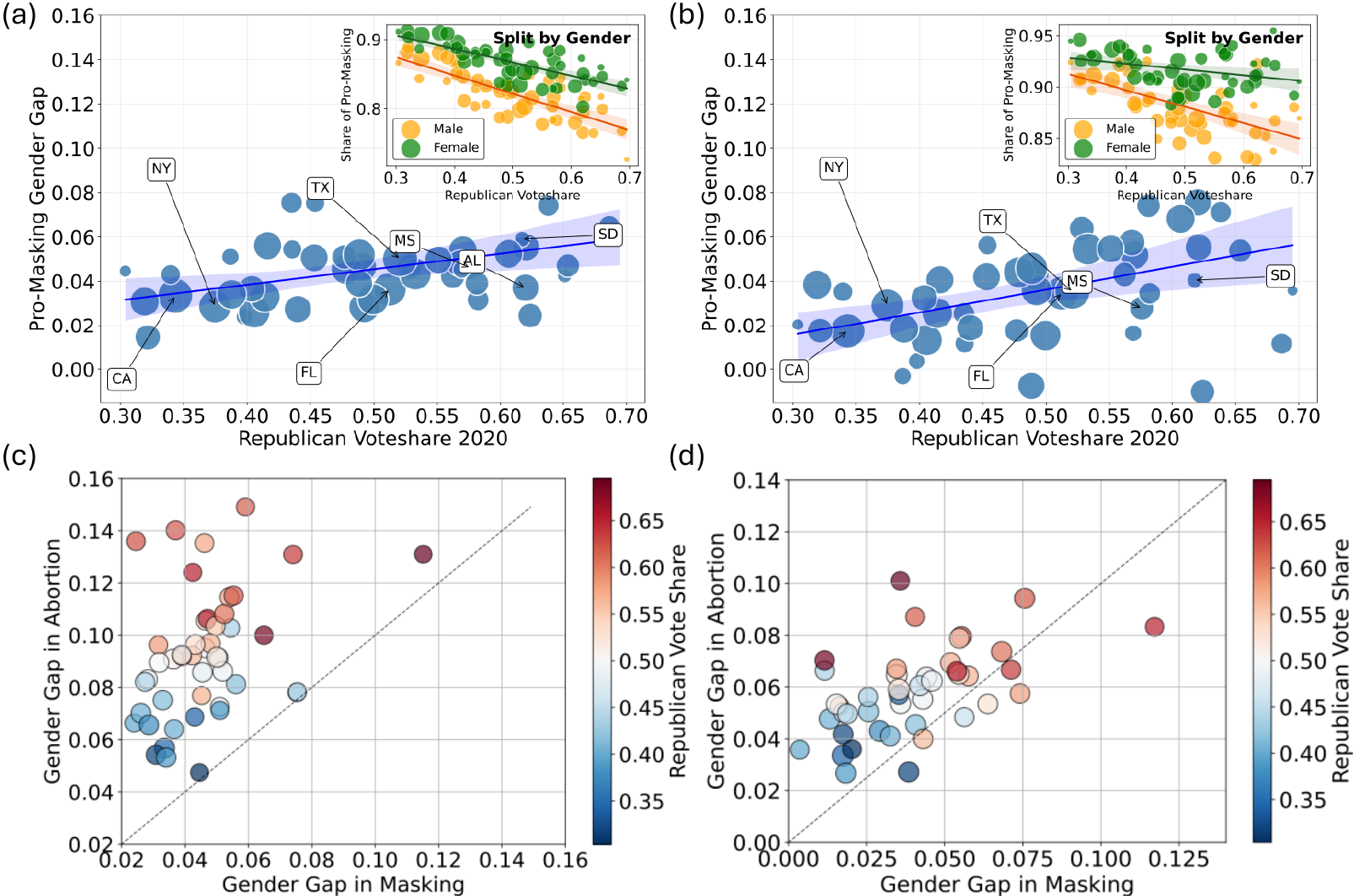}
    \caption{\textbf{Gender Gap in Pro-Masking Attitudes on Twitter.} The gender gap in support for masking, calculated as the difference in the share of pro-masking tweets from women and men in a given state,increases with state's Republican vote share. The magnitude of this gap is smaller than  the gender gap observed for abortion attitudes. The inset displays the proportion of pro-masking tweets authored by men and women across states. Panel (a) shows the gender gap in pro-masking attitudes among all users (Pearson $r = 0.43$), while panel (b) restricts the analysis to users who also engaged in abortion-related discourse (Pearson $r = 0.46$). (c) Compares the gender gap in pro-masking attitudes to the gender gap in abortion at the state-level for all users. (d) Compares the gender gap in pro-masking attitudes to the gender gap in abortion at the state-level for users active in both abortion and masking discussions.}
    \label{fig:masking_gap}
\end{figure}

Using the same method to infer user gender described in this paper, we calculate the gender gap in pro-masking attitudes as the difference in the share of pro-masking tweets from women and men in a given state (Figure~\ref{fig:masking_gap}). While the gap  is positively associated with Republican vote share in the 2020 U.S. Presidential election, it is notably smaller than that observed for abortion. This pattern holds both for the full set of users (Pearson correlation $r=0.43$, Fig.~\ref{fig:masking_gap}(a)) and for the subset of approximately 262K users who participated in both abortion- and masking-related discussions ($r=0.46$, Fig.~\ref{fig:masking_gap}(b)). The correlation is substantially lower than the corresponding correlation of 0.81 observed in abortion-related discourse.

\begin{table}[!ht]
\centering
\scriptsize
\begin{tabular}{lcccccc}
\toprule
\textbf{Issue} & \textbf{Sample} & \textbf{\# Tweets} & \textbf{\# Users} & \textbf{Median Gap} & \textbf{Mean Gap} & \textbf{SD} \\
\midrule
\multirow{2}{*}{Abortion} 
  & All Users           & 9,982,073     & 805,382       & $0.093$  & $0.091$  & $0.022$ \\
  & Overlapping Users   & 5,234,493     & 262,852    & $0.061$  & $0.056$ & $0.023$ \\
\midrule
\multirow{2}{*}{Masking} 
  & All Users           & 7,928,447   & 1,987,570       & $0.046$  & $0.035$  & $0.016$ \\
  & Overlapping Users   & 2,010,762     & 262,852    & $0.036$  & $0.030$ & $0.023$ \\
\bottomrule
\end{tabular}
\caption{Comparing Gender Gap in Pro-Abortion and Pro-Masking Attitudes}
\label{tab:gender_gap_summary}
\end{table}

Figures~\ref{fig:masking_gap}(c) and (d) present a state-level comparison between the gender gap in pro-masking and pro-abortion attitudes, separately for all users and for users who were active in both masking- and abortion-related discourse. Across most states, the gender gap is notably larger in abortion attitudes. Table~\ref{tab:gender_gap_summary} provides a summary of the average gender gaps observed in both topics. Gender gaps in pro-abortion and pro-masking attitudes for each state is detailed in Table \ref{tab:gender_gap_state_tab}.

\begin{longtable}[!b]{lcccc}
\toprule
\textbf{State} & \textbf{Pro-Abortion (All)} & \textbf{Pro-Masking (All)} & \textbf{Pro-Abortion (Common)} & \textbf{Pro-Masking (Common)} \\
\midrule
\endfirsthead

\toprule
\textbf{State} & \textbf{Pro-Abortion (All)} & \textbf{Pro-Masking (All)} & \textbf{Pro-Abortion (Common)} & \textbf{Pro-Masking (Common)} \\
\midrule
\endhead

\endlastfoot

ALABAMA         & 0.140 & 0.037 & 0.094 & 0.076 \\
ALASKA          & 0.109 & 0.073 & 0.077 & 0.051 \\
ARIZONA         & 0.095 & 0.046 & 0.055 & 0.043 \\
ARKANSAS        & 0.136 & 0.025 & 0.073 & -0.010 \\
CALIFORNIA      & 0.057 & 0.034 & 0.034 & 0.017 \\
COLORADO        & 0.081 & 0.056 & 0.045 & 0.041 \\
CONNECTICUT     & 0.066 & 0.045 & 0.032 & 0.015 \\
DELAWARE        & 0.066 & 0.024 & 0.036 & 0.004 \\
FLORIDA         & 0.091 & 0.036 & 0.065 & 0.035 \\
GEORGIA         & 0.083 & 0.028 & 0.054 & 0.036 \\
HAWAII          & 0.069 & 0.043 & 0.057 & 0.035 \\
IDAHO           & 0.131 & 0.074 & 0.067 & 0.071 \\
ILLINOIS        & 0.070 & 0.026 & 0.048 & 0.013 \\
INDIANA         & 0.115 & 0.054 & 0.069 & 0.052 \\
IOWA            & 0.106 & 0.046 & 0.053 & 0.064 \\
KANSAS          & 0.092 & 0.042 & 0.040 & 0.043 \\
KENTUCKY        & 0.115 & 0.055 & 0.080 & 0.055 \\
LOUISIANA       & 0.080 & 0.060 & 0.050 & 0.054 \\
MAINE           & 0.103 & 0.054 & 0.066 & 0.012 \\
MARYLAND        & 0.066 & 0.014 & 0.042 & 0.018 \\
MASSACHUSETTS   & 0.054 & 0.031 & 0.027 & 0.038 \\
MICHIGAN        & 0.086 & 0.046 & 0.064 & 0.044 \\
MINNESOTA       & 0.091 & 0.051 & 0.061 & 0.042 \\
MISSISSIPPI     & 0.135 & 0.046 & 0.172 & 0.028 \\
MISSOURI        & 0.097 & 0.048 & 0.064 & 0.058 \\
MONTANA         & 0.077 & 0.045 & 0.053 & 0.016 \\
NEBRASKA        & 0.096 & 0.032 & 0.067 & 0.035 \\
NEVADA          & 0.073 & 0.051 & 0.050 & 0.018 \\
NEW HAMPSHIRE   & 0.078 & 0.075 & 0.048 & 0.056 \\
NEW JERSEY      & 0.075 & 0.033 & 0.050 & 0.026 \\
NEW MEXICO      & 0.078 & 0.075 & 0.056 & 0.025 \\
NEW YORK        & 0.066 & 0.029 & 0.043 & 0.029 \\
NORTH CAROLINA  & 0.090 & 0.032 & 0.053 & 0.015 \\
NORTH DAKOTA    & 0.124 & 0.042 & 0.083 & 0.117 \\
OHIO            & 0.096 & 0.043 & 0.065 & 0.055 \\
OKLAHOMA        & 0.106 & 0.047 & 0.066 & 0.054 \\
OREGON          & 0.064 & 0.037 & 0.041 & 0.033 \\
PENNSYLVANIA    & 0.086 & 0.052 & 0.062 & 0.046 \\
RHODE ISLAND    & 0.071 & 0.051 & 0.017 & -0.003 \\
SOUTH CAROLINA  & 0.104 & 0.050 & 0.079 & 0.055 \\
SOUTH DAKOTA    & 0.149 & 0.059 & 0.087 & 0.041 \\
TENNESSEE       & 0.108 & 0.052 & 0.074 & 0.068 \\
TEXAS           & 0.092 & 0.050 & 0.059 & 0.035 \\
UTAH            & 0.092 & 0.039 & 0.058 & 0.074 \\
VERMONT         & 0.047 & 0.045 & 0.036 & 0.020 \\
VIRGINIA        & 0.082 & 0.027 & 0.050 & 0.019 \\
WASHINGTON      & 0.053 & 0.034 & 0.027 & 0.018 \\
WEST VIRGINIA   & 0.100 & 0.065 & 0.070 & 0.012 \\
WISCONSIN       & 0.092 & 0.039 & 0.053 & -0.007 \\
WYOMING         & 0.131 & 0.115 & 0.101 & 0.036 \\

\caption{Gender gap in pro-abortion and pro-masking attitudes by state, among all users and users active in both issue discourses.}
\label{tab:gender_gap_state_tab}
\end{longtable}
\end{document}